# A Comprehensive Insights into Drones: History, Classification, Architecture, Navigation, Applications, Challenges, and Future Trends

[1]Ruchita Singh (Corresponding Author) , [2]Sandeep Kumar

[1,2]*Bharat Electronics Limited , Ghaziabad, Uttar Pradesh 201010, India,*

[1] ruchita@bel.co.in, [2]sandeepkumar@bel.co.in

DECLARATION: THIS RESEARCH RECEIVED NO SPECIFIC GRANT FROM ANY FUNDING AGENCY IN THE PUBLIC, COMMERCIAL, OR NOT-FOR-PROFIT SECTORS.

*Abstract*— **Unmanned Aerial Vehicles (UAVs), commonly known as Drones, are one of 21st century most transformative technologies. Emerging first for military use, advancements in materials, electronics, and software have catapulted drones into multipurpose tools for a wide range of industries. In this paper, we have covered the history, taxonomy, architecture, navigation systems and branched activities for the same. It explores important future trends like autonomous navigation, AI integration, and obstacle avoidance systems, emphasizing how they contribute to improving the efficiency and versatility of drones. It also looks at the major challenges—technical, environmental, economic, regulatory and ethical—that limit the actual take-up of drones, as well as trends that are likely to mitigate these obstacles in the future. This work offers a structured synthesis of existing studies and perspectives that enable insights about how drones will transform agriculture, logistics, healthcare, disaster management, and other areas, while also identifying new opportunities for innovation and development.**

*Index Term*—**UAV, Drone, Architecture, Classification, Navigation System, Applications, Challenges**

## 1. INTRODUCTION

Unmanned Aerial Vehicles or UAVs popularly known as Drones have emerged as one of the most potent technologies of the 21st century. The history of UAVs (unmanned aerial vehicles) dates back as early as the 1900s, with the development of simple radio-controlled aircraft that were used for military purposes. Breakthroughs in materials, electronics and software in recent decades have turned drones from specialized military tool to ubiquitous platforms that can be used for a wide range of applications. Now drones are deployed for everything from surveillance and logistics to environmental monitoring and disaster response. Generally, Drones are classified based on their size, Arduino functionality, and operational ranges, showcasing the versatility of this technology. The fact that drone applications also vary significantly due to diverse industries and needs, can also be noticed in categories like nano drones, micro drones, small drones, tactical drones and large UAVs that are used for high-altitude and long-endurance missions.

Comprising an intricate amalgam of mechanical, electronic, and software components, the architecture of drones is the backbone of their formidable flight capabilities, allowing them to soar through the skies with a grace and accuracy that mere vehicles could never replicate. Modern drones are often a lightweight frame to which propulsion systems, batteries, sensors, and a control unit or several other components are attached. These components provide the foundation for navigation systems that utilize technologies such as Global Positioning System (GPS), Inertial Navigation Systems (INS), and more recently, Artificial Intelligence (AI)-based algorithms. These technologies allow drones to function autonomously or semi-autonomously in complex environments and leads us to the next step in the evolution of navigation systems. Previously, manual manipulation and GPS data have been the mainstays of models, whereas today's system utilise AI, computer vision and various high-tech sensing technology to navigate which improves the obstacle avoidance system of drones in real time and optimize the route, allowing them to make autonomous decisions with less reliance on individuals.

Drones are being used increasingly by all industries due to its endless versatility and efficiency they. For instance, in agriculture, drones enable precision farming and crop monitoring in construction, drones help with site inspections and progress monitoring. Logistics has been transformed by drones delivering last-mile shipments, for example, while public safety, environmental conservation, and media production, among others, have all taken advantage of the special capabilities of drones. Specific examples of their adoptive transformative nature, such as their use in disaster-response search-and-rescue missions or in delivering medical goods to inaccessible locations via messengers, also abound.

Nonetheless, as drones become more essential, they have a-number-of obstacles to overcome that may stand in the way of their growth and adoption. There are still significant barriers to the technology in terms of regulatory hurdles, cybersecurity vulnerabilities and technical limitations like battery life and payload capacity. The impressive pace of advancement in this field also raises ethical and societal issues, especially around privacy and misuse. This research paper covers topics such as the historical development, taxonomy, architectural structures, navigation techniques, applications, and studying of interest as well as what future trends, challenges, and opportunities might arise. Through this analysis, the paper seeks to shed light on the motivations behind the growing prominence of drones and their potential to shape the future in various fields.

*1.1 Motivation*

We are motivated to present a survey concerning the state of the art of UAV development and applications due to the identification of significant gaps in comparing more than 50 survey papers about drones covering different relevant fields of research. Ref. Table 1.for the analysis of papers to identify the gap for survey. Although multiple existing studies discuss certain aspects of the field such as navigation, architecture or applications, very few provide an organization of all existing relevant content in this domain, there exist no review that discusses in one read history of the field, its classification, its architecture, the navigation and control, challenges, future trend, and applications, if any absolutely. This Survey aims to ameliorate these deficiencies and offer a useful resource to researchers, industry players, and policy makers wishing for a more comprehensive understanding, and authoritative synthesis of the topic, to identify avenues for future work and innovation.

*Table 1: Analysis of papers to identify the gap for survey.*

| Paper Ref. | History | Classification | Architecture | Navigation & Control | Application | Case Study | Challenges |
|---|---|---|---|---|---|---|---|
| [1] | ✓ | ✓ | ✗ | ✓ | ✗ | ✗ | ✗ |
| [2] | ✗ | ✗ | ✓ | ✗ | ✓ | ✗ | ✗ |
| [3] | ✓ | ✓ | ✓ | ✓ | ✓ | ✓ | ✗ |
| [4] | ✓ | ✓ | ✓ | ✗ | ✓ | ✗ | ✗ |
| [5] | ✗ | ✗ | ✗ | ✗ | ✓ | ✗ | ✗ |
| [6] | ✗ | ✓ | ✗ | ✗ | ✓ | ✗ | ✗ |
| [7] | ✓ | ✓ | ✓ | ✗ | ✓ | ✗ | ✓ |
| [8] | ✓ | ✓ | ✗ | ✓ | ✓ | ✗ | ✓ |
| [9] | ✗ | ✗ | ✗ | ✓ | ✗ | ✗ | ✓ |
| [10] | ✗ | ✗ | ✓ | ✗ | ✗ | ✗ | ✗ |
| [11] | ✗ | ✓ | ✗ | ✗ | ✓ | ✓ | ✗ |
| [12] | ✗ | ✓ | ✗ | ✓ | ✓ | ✓ | ✓ |
| [13] | ✗ | ✗ | ✗ | ✓ | ✓ | ✗ | ✗ |
| [14] | ✗ | ✗ | ✗ | ✓ | ✓ | ✓ | ✗ |
| [15][16][17][18] | ✓ | ✗ | ✗ | ✗ | ✗ | ✗ | ✗ |
| [19] | ✗ | ✗ | ✗ | ✓ | ✗ | ✗ | ✗ |
| [20] | ✗ | ✗ | ✗ | ✗ | ✓ | ✗ | ✗ |
| [21] | ✗ | ✗ | ✗ | ✓ | ✓ | ✗ | ✗ |
| [22] | ✗ | ✗ | ✗ | ✓ | ✓ | ✗ | ✗ |
| [23] | ✗ | ✗ | ✓ | ✓ | ✓ | ✗ | ✗ |
| [24] | ✓ | ✗ | ✗ | ✓ | ✓ | ✗ | ✓ |
| [25] | ✗ | ✗ | ✗ | ✓ | ✗ | ✗ | ✓ |
| [26] | ✗ | ✗ | ✗ | ✓ | ✓ | ✓ | ✓ |
| [27] | ✗ | ✓ | ✗ | ✓ | ✓ | ✗ | ✓ |
| [28][29] | ✗ | ✓ | ✗ | ✓ | ✓ | ✗ | ✓ |
| [30] | ✗ | ✗ | ✗ | ✗ | ✓ | ✓ | ✓ |
| [31] | ✓ | ✓ | ✗ | ✗ | ✓ | ✓ | ✓ |
| [32] | ✓ | ✓ | ✗ | ✗ | ✓ | ✓ | ✓ |
| [33] | ✗ | ✗ | ✗ | ✗ | ✓ | ✓ | ✓ |
| [34] | ✗ | ✓ | ✓ | ✗ | ✓ | ✗ | ✓ |
| [35] | ✗ | ✗ | ✗ | ✓ | ✗ | ✗ | ✗ |
| [36][37] [38][39] | ✗ | ✗ | ✗ | ✓ | ✓ | ✓ | ✗ |
| [40] | ✗ | ✗ | ✗ | ✓ | ✓ | ✓ | ✗ |
| [41] | ✗ | ✗ | ✗ | ✓ | ✓ | ✗ | ✗ |
| [42] | ✗ | ✗ | ✗ | ✓ | ✗ | ✗ | ✗ |
| [43] | ✗ | ✗ | ✗ | ✓ | ✗ | ✗ | ✓ |
| [44] | ✗ | ✗ | ✗ | ✗ | ✓ | ✗ | ✓ |
| [45] | ✗ | ✗ | ✗ | ✓ | ✓ | ✗ | ✓ |
| [46] | ✗ | ✗ | ✗ | ✓ | ✗ | ✓ | ✓ |
| [47] | ✗ | ✗ | ✗ | ✗ | ✗ | ✓ | ✓ |
| [48] | ✓ | ✓ | ✓ | ✓ | ✗ | ✓ | ✗ |
| [49] | ✗ | ✓ | ✗ | ✗ | ✓ | ✓ | ✓ |
| [50] | ✗ | ✗ | ✓ | ✗ | ✗ | ✗ | ✗ |
| [51] | ✓ | ✗ | ✗ | ✗ | ✓ | ✗ | ✗ |
| [52] | ✓ | ✗ | ✗ | ✗ | ✗ | ✗ | ✗ |
| [53] | ✗ | ✗ | ✗ | ✗ | ✗ | ✗ | ✗ |
| [54] | ✗ | ✗ | ✗ | ✗ | ✗ | ✗ | ✗ |
| [55] | ✗ | ✓ | ✓ | ✗ | ✗ | ✗ | ✗ |
| [56] | ✗ | ✓ | ✓ | ✗ | ✗ | ✗ | ✓ |
| This Survey | ✓ | ✓ | ✓ | ✓ | ✓ | ✓ | ✓ |

*1.2 Prisma Analysis of Survey Paper*

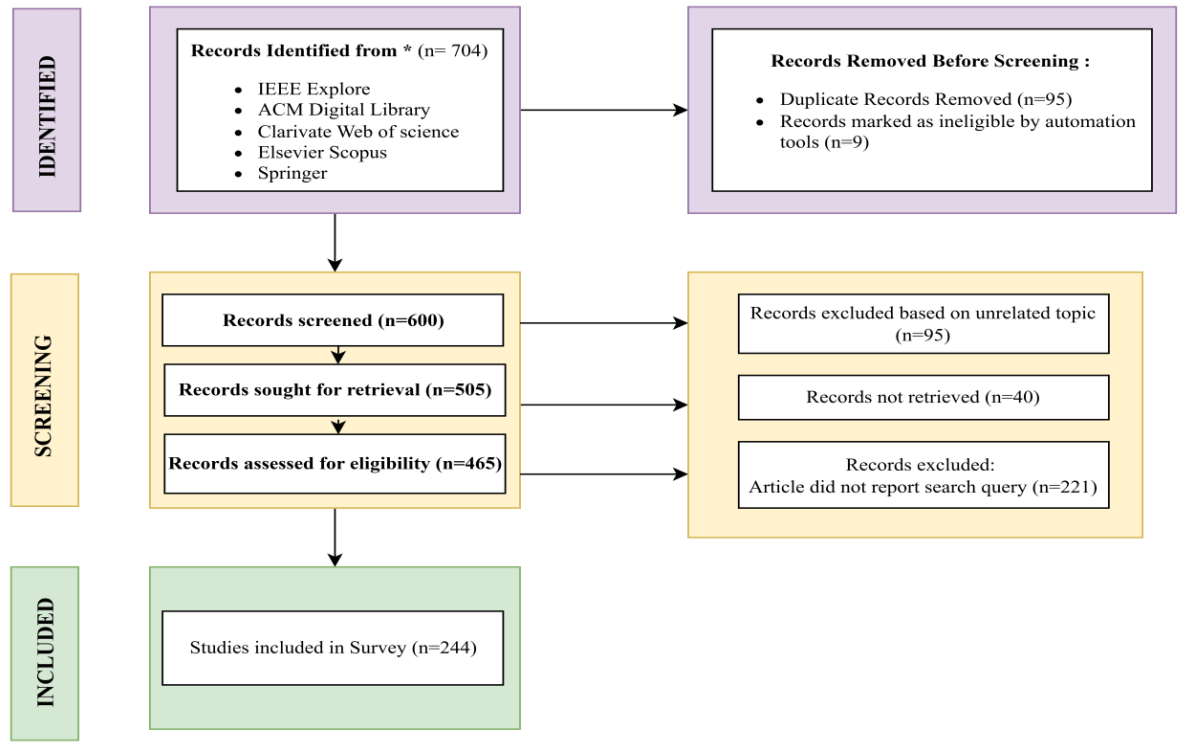

*Figure 1: PRISMA inclusion/exclusion criteria diagram. Exclusion reasons are presented in columns on the right side, Resulting in the 244 papers for review.*

The study follows the Preferred Reporting Items for Systematic Reviews and Meta-Analyses (PRISMA) guidelines [57], which outline three key stages: identification, screening, and inclusion. Each stage is explicitly defined and illustrated in Figure 1.

*1.2.1 Inclusion Criteria:*

Reviews focusing on UAV usage and technologies across various domains (history, classification, navigation, applications etc.) were considered.

*1.2 2 Exclusion Criteria:*

Non-secondary studies were excluded, Studies where UAV did not stand for Unmanned Aerial Vehicles were omitted, Articles not written in English were excluded, Studies lacking a defined search process, such as unreported search queries or database details, were omitted, Papers that were inaccessible were excluded.

A critical exclusion criterion was the absence of a "defined search process," directly inspired by [58]. This required that search queries and databases be explicitly detailed in the article through text or figures. Applying these criteria resulted in the inclusion of 244 systematic literature review (SLR) articles as shown in Fig.1 of PRISMA analysis.

*1.3 Organization of paper*

This paper is structured into nine main sections as illustrated in Figure 2. Here Drone Survey, starts with a light introduction including the motivation of this survey. It also addresses the implementation of the PRISMA methodology for conducting a systematic review and structuring the papers. This is followed by a second section that gives detail about drone evolution, and history. Third Section of paper classifies the drone based on various parameters such as size, weight etc.

*Table 2: Abbreviations*

| | |
|---|---|
| **UAV** | Unmanned Aerial Vehicles |
| **GPS** | Global Positioning System |
| **INS** | Inertial Navigation Systems |
| **AI** | Artificial Intelligence |
| **PRISMA** | Systematic Reviews and Meta-Analyses |
| **SLR** | Systematic literature review |
| **PTA** | Pilotless Target Aircraft |
| **VTOL** | vertical takeoff and landing |
| **IMU** | Inertial Measurement Unit |
| **PID** | Proportional-Integral-Derivative |
| **RRT** | Rapidly exploring Random Trees |
| **INS** | Inertial Navigation System |
| **SLAM** | Simultaneous Localization and Mapping |
| **GNSS** | Global Navigation Satellite Systems |
| **SRM** | Switched Reluctance Motors |
| **LiDAR** | Light Detection and Ranging |
| **HAP** | High Altitude Platforms |
| **MAP** | Medium Altitude Platforms |
| **LAP** | Low Altitude Platforms |

To cover up the fourth part, we will discuss in detail about the architecture of drones which includes their hardware and software components like propulsion system, communication module, sensors and other subsystems. The fifth section focuses on the mechanisms for navigation and control on drones, including technologies such as GPS, AI-based path-planning etc. In the sixth section, we discuss the various sectors utilizing drones, such as military, agriculture, logistics, healthcare, and disaster management. Section seven discusses the key challenges and Future Trends. In the eighth part, a case study, providing practical insight into the implementation and efficiency of UAVs in actual situations. In the end, the paper concludes with summarised findings and re-highlighting the significance of drones and suggesting avenues for future research.

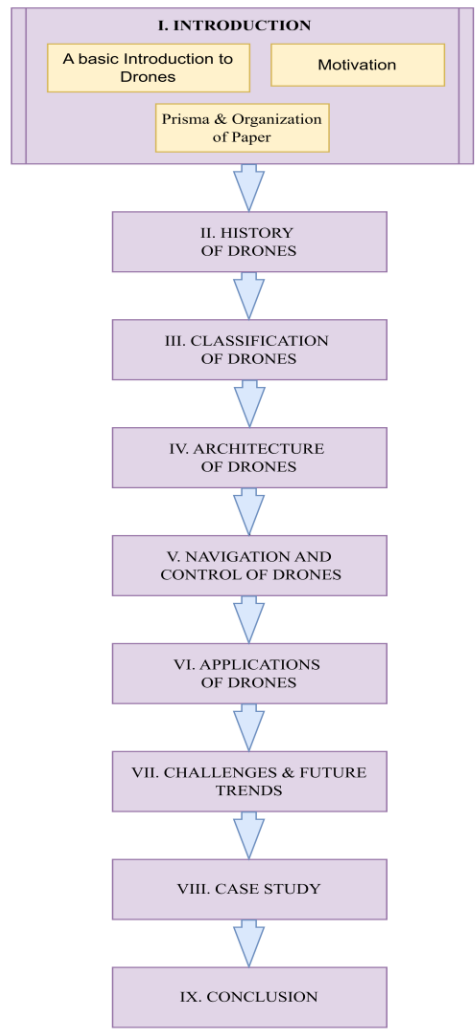

*Figure 2: Organization chart for Survey Paper*

## 2. History of Drones

Unmanned Aerial Vehicles (UAVs) have emerged as significant outcomes of technological advancements driven by military conflicts. Unmanned balloons were employed in the middle of the 18th century for offensive purposes, these types of aircraft or Unmanned ballons had first appeared during World War I and this marks the starting point for the UAV development [59]. Table 3 gives a brief history of drone's year wise and Figure 3. gives breakthrough in drones in each era.

## 2.1 WORLD WAR ERA.

Unmanned Aerial Vehicles (UAVs) had a long history of development, the earliest of which, was Elmer Sperry's prototype with an autopilot. In 1917, it was tested using modified Curtis N-9s, which were able to fly automatically with a limited degree of accuracy. Kettering Bug, which was finished in November 1917, further advanced the UAV concept but ultimately remained unused in combat due to the conclusion of World War I.

The post-war period saw the creation of Pilotless Target Aircraft (PTA) like the Queen Bee in the UK. It was the first UAV to complete missions and return unless destroyed, earning the nick name "drone"- making it the first, or deviant, drone in the aviation history.

UAV developments were fast tracked with the onset of World War II. Improvements in design and functionality were placed in production with the US versions of Reginald Denny's Radio Plane series, particularly ending in the OQ-14, some of which were still in service long after the war ended. On the other hand, the German V-1 introduced in 1944, was designed to carry an 850 kg warhead to a range of 320 km, which demonstrates the increasing strategic value of UAV technology [60][61][62][63].

## 2.2 COLD WAR ERA

The 1950s marked a turning point in UAV development with the pursuit of supersonic pilotless target aircraft. In 1953, Northrop's Radio-plane division developed the AQM-35, which achieved Mach 1.55 in 1956. Designed to train air defence missile units, the program was discontinued as the UAV's speed exceeded tracking capabilities.[64]

The vulnerabilities of manned reconnaissance, highlighted by the 1960 U-2 incident over the Soviet Union, spurred the shift to unmanned systems. The high risks and costs of the U-2 program prompted the development of safer, cost-effective UAV alternatives for intelligence missions.[65]

In 1960, the U.S. Air Force tasked Ryan Aeronautical with adapting the Ryan Model 147 into reconnaissance drones, resulting in the BQM-34 Firebee, or "Lightning Bug." These drones proved invaluable during the Vietnam War, advancing aerial reconnaissance and, by 1972, enabling live data transmission to ground stations, solidifying their role in modern military operations. [60][61][62][63][64].

## 2.3 POST-VIETNAM WAR ERA

Israel would become a leader in UAV development during the 1970s, producing successful models such as the Mastiff and Scout. But by the 1980s, the U.S. led the world in UAV production.

In 1991, during Operation Desert Shield in Iraq, the United States and its allies achieved a quick victory using advanced technology. In the Yugoslav War, reconnaissance was, once again, highlighted (NATO and UN intervened in 1995, with Operation Deliberate Force) as an important task in military planning. In the case of UAVs like the Pioneers and Predator, operating in any weather with SAR systems and satellite data links was invaluable. By then 15% of US reconnaissance aircraft were UAVs and land forces were deploying low-cost, hand-launched drones for use in the field [60][61][62][63][66].

## 2.4 21ST CENTURY

In the modern era, stealth capability and the integration of onboard systems into a unified, complex unit are essential features of fighter aircraft. This takes us a step closer toward an information warfare, which seeks informational superiority over the enemy This evolution of the MQ-9 Reaper UAV represents a modern day strike weapon system, the bigger, stronger successor to the Predator. Built for long-endurance, high-altitude surveillance and hunter-killer missions.
.

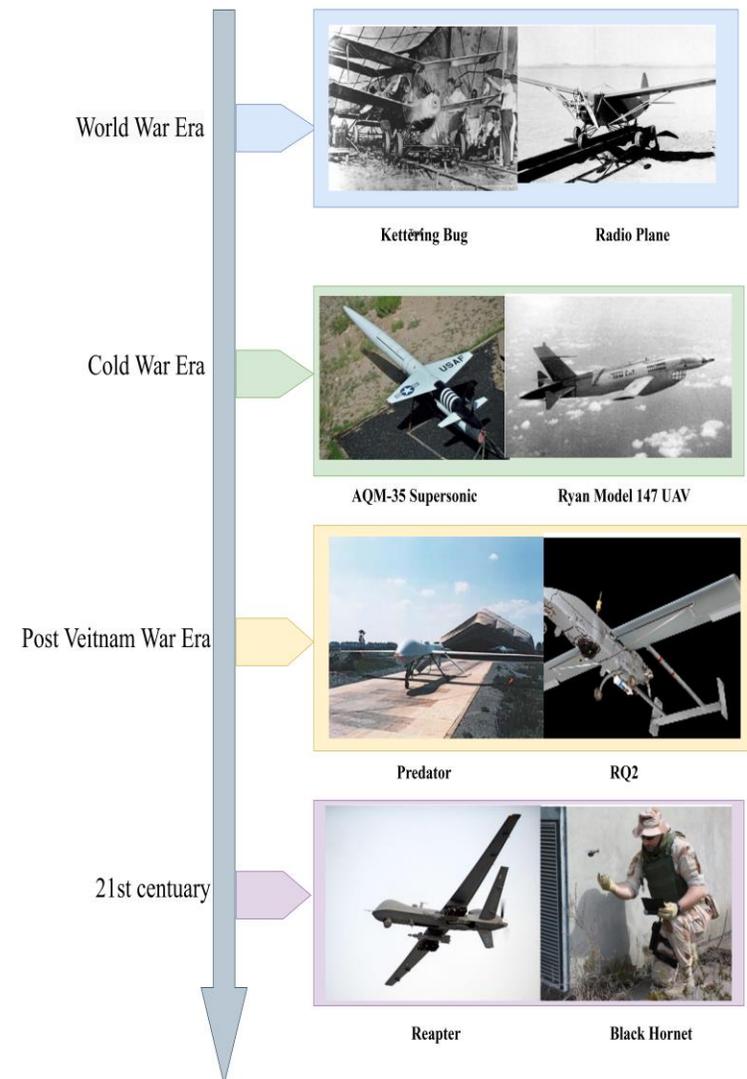

Figure 3: A Brief History of Drones [67]

In asymmetric warfare, small, low-cost, hand-launched UAVs have become invaluable, particularly for the army and marines. Drones such as the Black Hornet Nano, are electric-powered, silent, portable, and controllable via laptops or tablets, making them highly effective in urban environments. Equipped with EO or IR cameras, drones transmit real-time images to operators. The Black Hornet Nano, used by Norwegian, British, and US forces, is a micro-UAV measuring 10×2.5 cm, weighing just 16 g, and can be operated after only 20 minutes of training [67].

## 2.5 HUNGARIAN UAV DEVELOPMENTS

Aero-Target Bt has manufactured the Meteor PTA family for the Hungarian Defence Forces (HDF) since 1999, focusing on air defence training. In 2005, a modernization effort produced the Meteor3, a low-cost, all-Hungarian target system. Moreover, the skylark I LE purchased from Israel in 2009 was the backbone of Hungary's aerial observation which enormously increased the surveillance capabilities of the country [67][68].

*Table 3: Major Achievements in Drone from 1917 to 2023*

| S. No | Year | Major-Achievement | Domain | Key Findings | Country of Origin |
|---|---|---|---|---|---|
| **1.** | 1917 [69] | First Drone Prototype (Kettering Bug) | Miliary | Early prototype of unmanned aerial torpedoes. Used in testing during World War I but never deployed in combat. | USA |
| **2.** | 1935 [70] | First Radio Controlled Drone (Radio OQ-2) | Miliary | First mass-produced UAV, used for target practice. Used in World War II for training anti-aircraft gunners. | USA |
| **3.** | 1960 [71] | First Reconnaissance Drone (Ryan-Firebee) | Military | The Ryan Fire bee was among the first widely used reconnaissance drones, developed during the Cold War. Used in Surveillance during the Vietnam War and for Cold War reconnaissance missions. | USA |
| **4.** | 1981 [72] | Israel's Scout Drone (Israel Aerospace Industries (IAI) Scout Drone) | Military | It was designed for real-time video surveillance and helped Israel achieve aerial superiority without risking pilot lives. Used In: Israeli Defense Forces (IDF) in the 1982 Lebanon War. | Israel |
| **5.** | 1980 [73] | GPS Drone (MQ-1) | Military | Used in Intelligence, surveillance, and reconnaissance (ISR) missions, particularly in the Balkans and Middle East. The Predator was widely used in conflicts like the Bosnian War and later in Afghanistan and Iraq. | USA |
| **6.** | 1997 [74] | First Commercial Drone (Yamaha RMAX) | Commercial, non-military applications, | The drone was designed primarily for agricultural spraying, offering precision in delivering fertilizers, herbicides, and pesticides over crops. Used in RMAX has been widely used in Japan's rice fields and other agricultural settings for crop dusting and spraying | JAPAN |
| **7.** | 1998 [75] | RQ-4 Global Hawk | Miliary | Long-endurance, high-altitude surveillance. Used in Intelligence and surveillance missions worldwide. | USA |
| **8.** | 2001 [76] | Predator Drone (MQ-1 Predator Drone) | Miliary | Pioneer in reconnaissance and later armed drones. Used in Middle East for surveillance and combat missions | USA |
| **9.** | 2010 [77] | First Consumer Drones (Parrot AR Drone) | Consumer, Entertainment | Early consumer drone with smartphone control. Used in Commercial drone market for recreational and photographic purposes. | France |
| **10.** | 2013 [78] | DJI Phantom Series | Consumer, Commercial | Revolutionized consumer drone photography. Used in Photography, videography, mapping, and inspections globally | China |
| **11.** | 2014 [79] | Amazon Delivery Drone Deliver (Amazon Prime Air) | Logistics, Delivery | First commercial drone delivery tests. Used In: Pilot programs for package deliveries. | USA |
| **12.** | 2016 [80] | DJI Mavic Series | Consumer, Professional | Portable, foldable drones for professional use. Used in Photography, mapping, search-and-rescue missions globally. | China |
| **13.** | 2020 [81] | Drone Racing and AI (Drone Racing League) RacerAI | commercial | The RacerAI drones were equipped with advanced AI systems that enabled them to fly autonomously, without human intervention, and navigate the racing course at high speeds. | USA |
| **14.** | 2018 [82] | 5G Drone Connectivity (5g Connected Drone) | Commercial | Used in infrastructure inspection, agriculture, and emergency response. | Korea, China |
| **15.** | 2019 [83] | Boeing MQ-25 Stingray | Military | First carrier-based refueling drone. Used in U.S. Navy carrier refueling operations. | USA |
| **16.** | 2019 [84] | Skydio 2 | Consumer, Professional | Advanced AI-powered autonomous drone. Used in Autonomous filming, search-and-rescue operations. | USA |
| **17.** | 2016 2020 [85][86] | Swarm Drones LOCUST, Shenzhen's 1,000 Drone | Navy, Military | Cooperative Behavior, Cost-Effectiveness, Tactical Advantages, Scalability, Autonomous Operation | USA, China |
| **18.** | 2023 [87] | AI- Driven Autonomous Drone (AI based Navigation) | Non-Military | Enhanced autonomy and efficiency in various applications. Improved obstacle detection reduces operational risks. AI adaptability allows drones to learn and optimize perf. | USA |

## 3. Classification of Drone

The classification of drones provides a structured framework to enhance understanding, regulation, and utilization. It aids in design and development by aligning specifications with applications, by facilitating targeted regulatory measures for safe and lawful operations and enables industry-specific applications for improved efficiency. Classification plays a critical role in risk assessment, ensuring safety in diverse environments, and supports academic research. Moreover, it encourages market segmentation, economic optimization and ethical deployment, particularly in military and environmental applications. classification ensures that drones are designed, deployed, and managed effectively across various domains. Various parameters for Classification of drone were discussed in [88]. In this paper we introduced a novel-classification categorization parameters, shown in Fig. 4. Below section defined them in detail.

### 3.1. Drone Classification based on Design

#### 3.1.1 Classification Based on Size

Drones may vary in size to fulfill various specific operational needs. For example, smaller drones like Nano and Micro are ideal for close-range reconnaissance, larger drones like Tactical and Strike are best suited to military-grade missions. Drones classified based on their size directly impacts maneuverability, payload, capacity and operational applications of drone. Size based classification are as follows. **Nano drones**, of size less than 250 mm, are highly compact and lightweight they are optimized for indoor surveillance and reconnaissance. Small size of these drones enables enhanced portability and discreet deployment in confined spaces. **Micro drones**, measuring between 250 mm and 500 mm, are slightly larger and often equipped with basic cameras and sensors, making them suitable for lightweight outdoor applications such as short-range monitoring and inspections. **Mini drones**, ranging from 500 mm to 2 m, are widely utilized in sectors like agriculture and search-and-rescue operations due to their moderate payload capacity, extended range, and adaptability to diverse environments. **Small drones**, sized between 2 m and 5 m, are designed for industrial-scale operations, including infrastructure inspections and long-duration surveillance missions, owing to their enhanced endurance and advanced sensor integration. **Tactical drones**, measuring 5 m to 10 m, are specialized for military applications, offering capabilities such as high-precision reconnaissance, target acquisition, and battlefield support. **Strike drones**, exceeding 10 m in size, are equipped with significant payload capacities and advanced avionics, enabling them to perform strategic operations, such as long-range surveillance, payload delivery, and precision strikes. This classification highlights the correlation between drone dimensions and their technical and operational functionalities. Table 4 and Fig. 5 Gives summarized details of classification. Above Size based classification were supported by studies from [91][92].

#### 3.1.2 Classification Based on Aerodynamics

Aerodynamic configurations influence the efficiency, flight range, and versatility of drones. The aerodynamic configuration determines adaptability in flight.

Table 4: Classification based on Size of Drone [93]

| S. No | Size | Dimensions | Example |
|-------|------|------------|---------|
| 1 | Nano | <250 mm | Black Hornet Nano |
| 2 | Micro | 250–500 mm | DJI Mavic Mini |
| 3 | Mini | 500 mm–2 m | Parrot ANAFI |
| 4 | Small | 2–5 m | AeroVironment Puma |
| 5 | Tactical | 5–10 m | MQ-1 Predator |
| 6 | Strike | >10 m | RQ-4 Global Hawk |

Drones can also be classified based on their aerodynamic configurations, which influence their operational efficiency and versatility. Table 5 and Fig. 5 gives summarized details of classification.

Fixed-wing drones are ideal for long-distance reconnaissance and mapping, offering superior endurance and energy efficiency for covering expansive areas. Fig. 6(c) show the types of fixed-wing classification. This type of classification study was supported by [94][95][96]. Flapping-wing drones mimic natural flight patterns, making them particularly effective for biomimetic and stealth operations where low visibility and noise reduction are critical. This study was supported by [97]. Rotary-wing drones provide exceptional maneuverability and hovering capabilities, making them well-suited for tasks such as inspections and aerial videography in confined or complex environments. Fig.6 (a) gives the sub classification of rotary wings. Lastly, hybrid drones [97] combine the vertical takeoff and landing (VTOL) capability of rotary-wing systems with the energy-efficient flight of fixed-wing designs, offering a versatile solution for missions requiring both precision and range. Fig 6(b) shows type of hybrid aerodynamic classification. Above classification information was concluded from [98][99].

Table 5: Classification of Drones based on Aerodynamics [100]

| S. No | Type | Example | Key Features |
|-------|------|---------|--------------|
| 1 | Fixed | SenseFly eBee | Efficient for long-distance missions |
| 2 | Flapping | RoboBee | Stealth and biomimetic design |
| 3 | Rotary | DJI Phantom 4 | Hovering and maneuverability |
| 4 | Hybrid | WingtraOne | Combines VTOL with long-endurance capabilities |

#### 3.1.3 Classification Based on Altitude

Drones can be categorized based on their operational altitude, which significantly impacts mission scope and feasibility. Fig 5 shows the altitude-based classification.

High Altitude Platforms (HAP) operate at altitudes above 20 km, making them ideal for atmospheric research, climate monitoring, and strategic reconnaissance due to their extended range and prolonged endurance [101]. Medium Altitude Platforms (MAP) function between 5 and 20 km, primarily used for surveillance, border patrol, and other mid-range operations requiring sustained observation. Low Altitude Platforms (LAP) operate below 5 km, well-suited for tasks such as infrastructure inspections, urban surveillance, and package delivery, where proximity to the ground is critical for accuracy and efficiency. Table 6 summarizes the details. Above inference were taken from [100][102].

Table 6: Classification of Drones based on Altitude [100]

| Altitude | Example | Operational Range | Application |
|----------|---------|-------------------|-------------|
| HAP | Zephyr S | >20 km | Atmospheric Research |
| MAP | MQ-9 Reaper | 5–20 km | Surveillance |

### 3.1.4 Classification Based on Wing Type

The type of wing configuration in drones significantly impacts their stability and operational efficiency. Multi-rotor drones are highly effective for short-distance, precise tasks, offering excellent maneuverability and the ability to hover in place, making them ideal for applications such as inspections and aerodynamics allow them to cover larger areas more videography. Table7 gives summary of wing type classification. Fig 5 shows the types of wing type classification.

Fixed-wing drone, on the other hand, are designed for high-efficiency, long range operations. Their effectively, making them suitable for tasks like surveying and mapping. This Wing type classification study were inferenced from [102][103][104].

Table 7: Classification of Drones based on Wing Types [102][103]

| Wing Type | Example | Applications |
|-----------|---------|--------------|
| Multi-Rotor | DJI Inspire 2 | Videography, Inspections |
| Fixed-Wing | Parrot Disco | Long-range missions |
| Single Rotor | Yamaha RMAX | Agricultural spraying |

### 3.1.5 Classification Based on Flying Mechanism

The flying mechanism of a drone plays a crucial role in determining its deployment capabilities and versatility

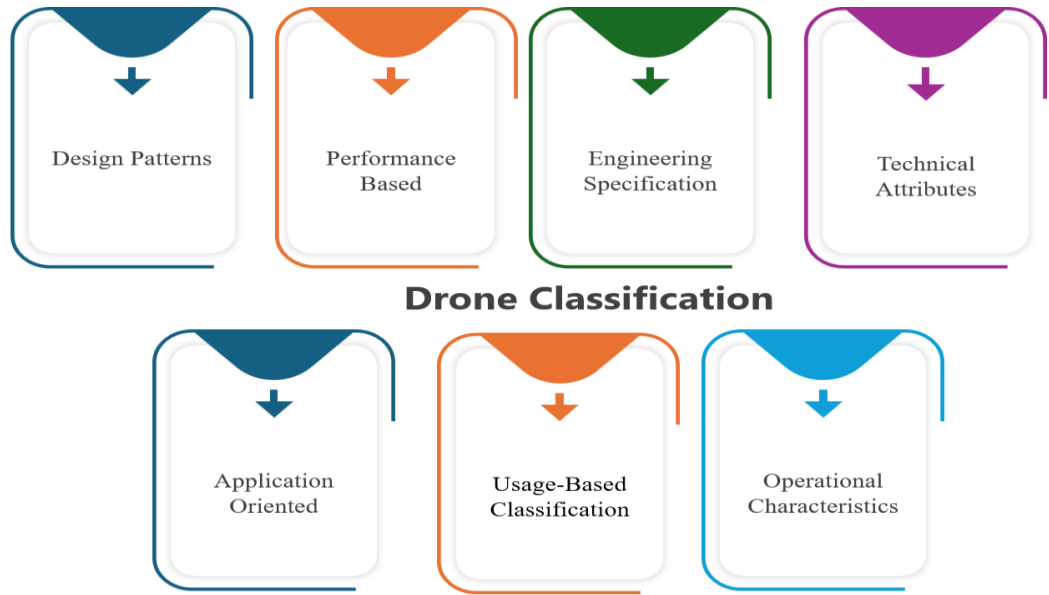

Figure 4: Major Categories of Drone Classification

Fig 5 gives types of flying mechanism-based classification. Fixed-wing drones are typically used for long-range reconnaissance and mapping, as their aerodynamic design allows for efficient, extended flight over large areas. Multi-copters provide exceptional stability and versatility, making them ideal for tasks such as inspections and videography, where precise control and the ability to hover are essential. Tilt-wing and tilt-rotor drones combine the best of both worlds, offering Vertical Take-Off and Landing (VTOL) capabilities for operation in confined spaces, while also enabling forward flight for greater range and speed, making them highly versatile for a variety of mission types. Table 8 summarizes this classification. Above information was supported by studies from [99][102][104][116].

Table 8: Classification of Drones based on Flying Mechanism [102]

| Flying Mechanism | Example | Applications |
|------------------|---------|--------------|
| Fixed Wing | SenseFly eBee | Mapping |
| Multi-copters | DJI Mavic 2 Pro | Inspections |
| Tilt-Wing/Tilt Rotor | Joby S4 | VTOL and long-distance versatility |

### 3.1.6 Classification Based on Payload

Payload capacity is a key factor in determining the type of sensor or cargo a drone can carry. Fig 5. Gives types of payload-based classification. Feather-weight drones (4 g–100 g) are equipped with small sensors or cameras, making them ideal for lightweight applications such as basic surveillance or close-range inspections. Lightweight drones (150 g–270 g) can carry basic payloads, often used for tasks like inspections or light surveying. Middle-weight drones (400 g–1460 g) can carry medium-sized sensors or small packages, offering a balance between payload capacity and maneuverability, making them suitable for a variety of industrial and commercial applications. Heavy-lift drones (greater than 1460 g) are designed to carry industrial-grade sensors or large cargo, enabling their use in more complex operations such as large-scale deliveries, mapping, or even construction tasks. Table 9. Gives summary of payload based classification. The Above Classification was supported by studies from [104].

Table 9: Classification of Drones based on Payload [104]

| Payload | Example | Capacity |
|---------|---------|----------|
| Feather Weight | RoboBee | <100 g |
| Light Weight | DJI Mini SE | 150–270 g |

| Middle Weight | DJI Mavic 3 | 400–1460 g |
| Heavy-Lift | Yuneec H920 Tornado | >1460 g |

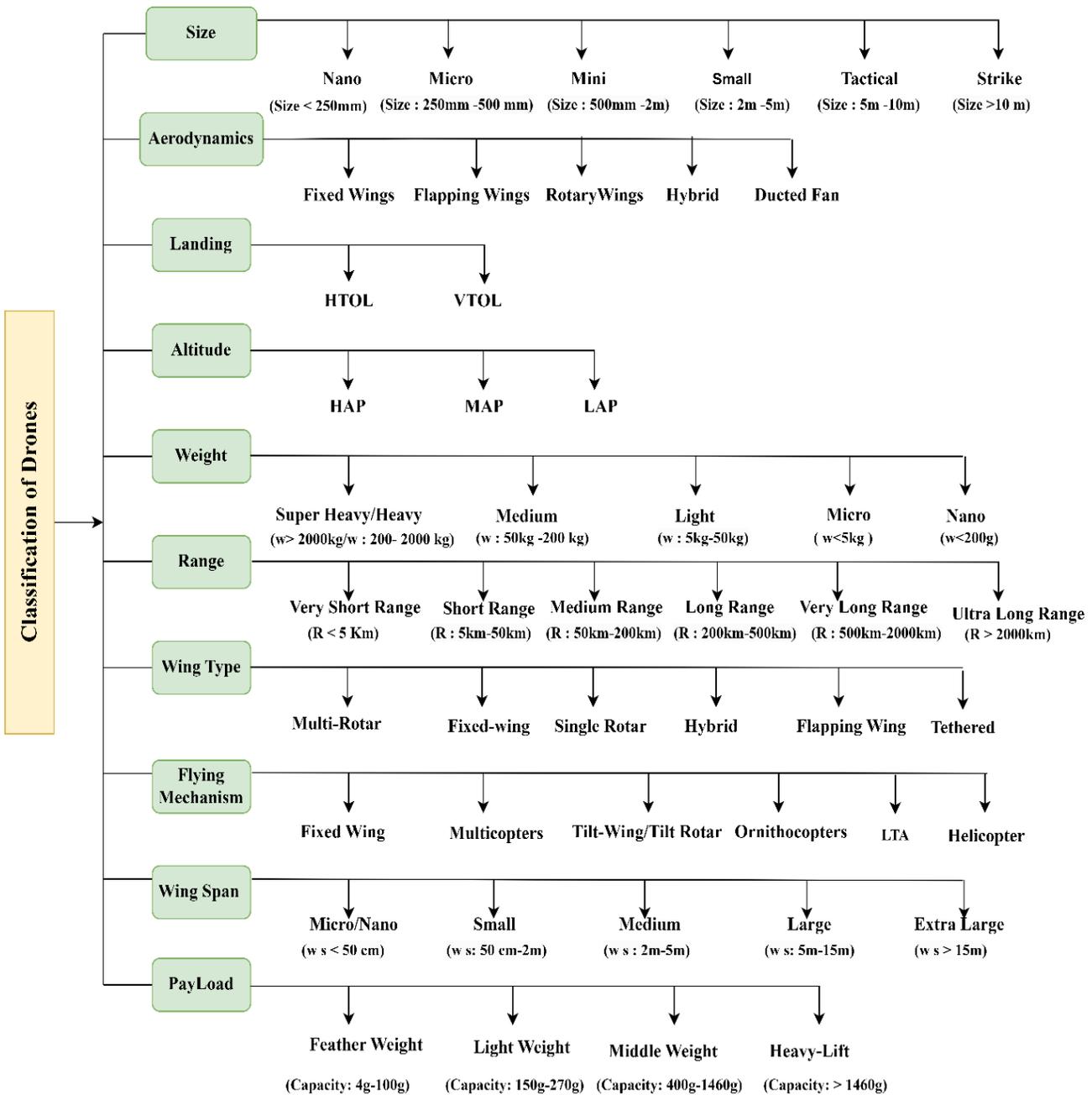

*Figure 5: Details Classification of Drone [99] [116]*

## 3.2 Drone Classification based on Performance

### 3.2.1 Classification Based on Power Source

Drones can also be categorized based on their power source, which influences their endurance, range, and operational capability. Battery-powered drones, like the DJI Mavic 3, are lightweight and ideal for small-scale applications, though they offer limited range. Fuel-powered drones, such as the Yamaha RMAX, provide longer endurance and can carry higher payloads. Solar-powered drones, like the Zephyr S, are capable of continuous flight, making them suitable for atmospheric research. Hydrogen fuel cell drones, such as the Doosan Mobility DS30, offer longer flight times and are environmentally friendly. Hybrid-powered drones combine battery and fuel, offering flexibility in operations. Above Classification information is supported by studies from [104].

### 3.2.2 Classification Based on Speed

Drones are also classified based on their speed. Low-speed drones, such as the DJI Mini 3, typically fly at speeds below

50 km/h and are suitable for photography and hobbyist applications. Medium-speed drones, like the DJI Matrice 300, operate at speeds between 50–150 km/h, making them ideal for industrial inspections. High-speed drones, such as jet-powered models, exceed speeds of 150 km/h and are employed in military reconnaissance. This information is supported by studies from [114][115].

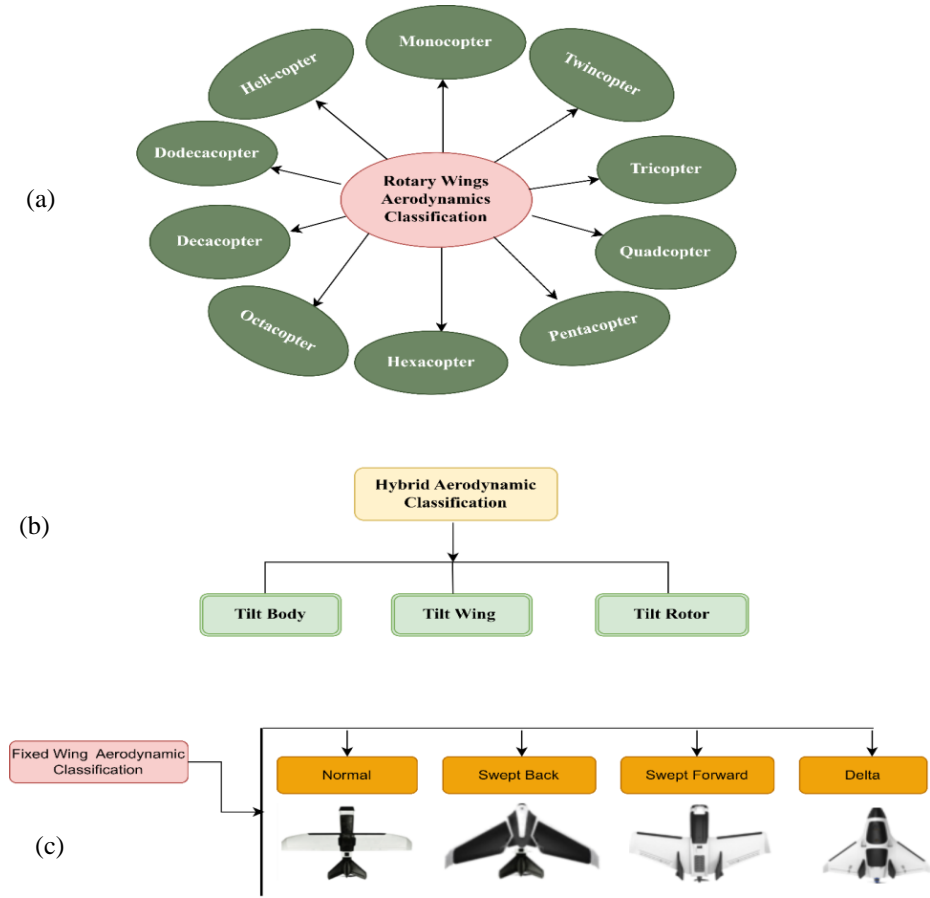

Figure 6: (a) Sub-Classification of Rotary Wings [107][108]. (b) Sub-Classification of Hybrid Aerodynamic Classification Type [99]. (c) Sub- classification of Fixed Wing Aerodynamic Classification Type [118].

### 3.2.3 Classification Based on Weight

Weight is a critical factor in drone classification, as it directly influences payload capacity, range, and endurance. Fig. 5 shows the type of weight-based classification. Nano drones, weighing less than 200 g, are ultra-light and designed for close-range operations, often used in surveillance or reconnaissance. Micro drones, weighing under 5 kg, are lightweight and portable, making them suitable for outdoor photography and basic monitoring tasks. Light drones, with a weight range of 5–50 kg, can carry heavier sensors and small cargo, enabling their use in applications such as environmental monitoring and precision agriculture. Medium drones, weighing between 50 and 200 kg, are equipped for industrial inspections and mid-range logistics, providing a balance between portability and capacity. At the upper end of the scale, super heavy drones, exceeding 2000 kg, are engineered for military-grade payloads and large cargo transport, supporting strategic operations and complex missions. [89] [90] discussed classification based on weight. Table 10 summarises the classification. This classification information is also supported by [109][100][102].

Table 10: Classification based on Weight [100][102]

| Weight | Example | Applications |
|---|---|---|
| **Nano** | Black Hornet Nano | Surveillance |
| **Micro** | DJI Mini 3 Pro | Photography |
| **Light** | SenseFly Albris | Industrial inspections |
| **Medium** | Bell APT 70 | Logistics |

### 3.2.4 Classification Based on Range

The operational range of a drone plays a significant role in determining its suitability for various missions. Fig. 5 shows the type of classification. Very short-range drones (less than 5 km) are typically used for indoor applications and short-range tasks, where the need for extended distance coverage is minimal. Short-range drones (5–50 km) are ideal for regional delivery services and inspections, allowing for efficient operations over a localized area. Medium-range drones (50–200 km) are commonly employed in agriculture and surveillance, offering sufficient coverage for monitoring large areas or agricultural fields. Long-range drones (200–500 km) are used for more extensive missions. This classification information is also supported by [100][102]. At the extreme end, ultra-long-range drones (greater than 2000 km) are designed for strategic surveillance missions, providing the capability to conduct long-duration operations over wide geographical areas, often for military or high-priority surveillance tasks. Table 11 shows summary of this classification.

Table 11: Classification based on Range [100][102]

| Range | Example | Distance |
|-------|---------|----------|
| **Very Short (<5 km)** | Tello Drone | Indoor applications |
| **Short (5–50 km)** | DJI Mini 3 Pro | Delivery |
| **Medium (50–200 km)** | AeroVironment Wasp AE | Surveillance |
| **Long (200–500 km)** | Hermes 900 | Border patrol |
| **Ultra Long (>2000 km)** | Global Hawk | Strategic missions |

### 3.3 Drone Classification based on Operational Characteristics

#### 3.3.1. Classification Based on Landing Mechanism

Landing mechanisms influence deployment in confined or open environments.

Drones can also be categorized based on their takeoff and landing mechanisms, which determine their deployment flexibility and operational environments. Horizontal Take-Off and Landing (HTOL) drones require a runway for operation, making them well-suited for long-endurance missions where efficient flight over extended distances is essential. In contrast, Vertical Take-Off and Landing (VTOL) drones can operate in confined spaces without the need for a runway, providing versatility for applications in urban environments, rugged terrains, or other areas with limited space for traditional takeoff and landing. Table 12. gives short summary of this classification. This classification information is also supported by [111][112][113].

Table 12: Classification based on Landing Mechanism [111]

| Landing Mechanism | Example | Features |
|-------------------|---------|----------|
| HTOL (Horizontal) | ScanEagle | Long endurance, runway required |
| VTOL (Vertical) | DJI Matrice 300 RTK | Confined-space operations |

#### 3.3.2 Classification Based on Communication and Control

The communication and control technology of drones determines their range and responsiveness.

Line-of-sight (LOS) drones, such as the DJI Mini SE, operate within a limited range, requiring direct visual control. Beyond visual line-of-sight (BVLOS) drones [117], like Amazon Prime Air, are used for extended-range delivery. Satellite-controlled drones, such as the Global Hawk, excel in long-distance military missions. 5G-enabled drones provide high-speed communication for real-time data exchange.

#### 3.3.3 Classification Based on Flight Environment

Drones can also be classified based on their flight environment.

Indoor drones, such as the Tello Drone, are compact and collision-resistant, optimized for indoor use. Outdoor drones, like the DJI Phantom 4, are designed for outdoor missions and are resistant to wind and equipped with GPS. Underwater drones, like the BlueROV2, are intended for aquatic environments. Space drones, such as NASA's Mars Helicopter, are adapted for low-gravity environments. This classification information is also supported by [99].

### 3.4 Technical Attributes

#### 3.4.1 Classification Based on Sensor Capability

The sensor capability of drones defines their mission-specific utility. Optical sensors, found in drones like the DJI Phantom 4, are used for aerial photography and object detection. Thermal sensors, like those on FLIR-equipped drones, are used for heat detection in search-and-rescue operations. LiDAR sensors, found on the DJI Matrice 300 RTK, enable terrain mapping and industrial inspections. Multispectral sensors, used in drones like the Parrot Bluegrass, are deployed for precision agriculture. Gas sensors in specialized industrial drones are used for air quality monitoring. This classification information is also supported by [112][115]

#### 3.4.2 Classification Based on Durability and Weather Resistance

Durability and weather resistance are also key classification factors. All-weather drones, such as the DJI Matrice 300 RTK, are built to withstand rain, wind, and extreme temperatures. Underwater drones, like the BlueROV2, are designed for submarine operations. Ruggedized drones, such as the Parrot Anafi USA, are used in military and disaster response scenarios. This classification information is also supported by [99].

### 3.5 Usage-Based Classification

Drones are classified based on their end-use, showcasing their versatility across various industries. Military drones such as the MQ-9 Reaper are primarily used for surveillance and target acquisition. Commercial drones, like the DJI Inspire 2, are widely employed for aerial photography and delivery tasks. Industrial drones, such as the SenseFly Albris, are commonly used for inspections and precision agriculture. Environmental drones like the Zephyr S are designed for climate monitoring and wildlife conservation, while medical drones, such as Zipline, are used for emergency medical supply delivery. Recreational drones, like the DJI Mini 3, are popular for aerial videography and hobby flying. This classification information is also supported by [116].

### 3.6 Classification Based on Autonomy Level/ Engineering Specifications

Drones can also be classified by their level of autonomy. Manual drones, like the Tello Drone, are fully operated by human controllers. Semi-autonomous drones, such as the DJI Mavic 2 Pro, combine automation with human intervention. Fully autonomous drones, like the Skydio 2, perform tasks independently of human control. This classification information is also supported by [116].

### 3.7 Summary Table of Drone classifications

Table 13. presents a novel approach to summarize the classification of drone by explicitly defining each classification criteria with specification, applications and classification basis.

*Table 13: Summary of Classification of Drones [88][89][90][91][92][93][94] ... [117][118]*

| Criteria | Categories | Example | Specifications | Applications/Features | Classification Basis |
|---|---|---|---|---|---|
| **Size** | Nano, Micro, Mini, Small, Tactical, Strike | Black Hornet, Global Hawk | Nano: <250 mm; Strike: >10 m | Nano: Close-range reconnaissance; Strike: Strategic surveillance and military operations | Design Parameters |
| **Aerodynamics** | Fixed, Flapping, Rotary, Hybrid | DJI Phantom, WingtraOne | Fixed Wing: Long flight range; Flapping: Biomimicry for stealth; Hybrid: VTOL & endurance | Mapping, stealth, hovering, endurance-based missions | Design Parameters |
| **Landing Mechanism** | HTOL (Horizontal), VTOL (Vertical) | ScanEagle, DJI Matrice 300 | HTOL: Needs runways; VTOL: Operates in confined spaces | HTOL: High endurance; VTOL: Urban delivery or reconnaissance in confined areas | Operational Characteristics |
| **Altitude** | HAP (High), MAP (Medium), LAP (Low) | Zephyr S, MQ-9 Reaper | HAP: >20 km; MAP: 5–20 km; LAP: <5 km | HAP: Atmospheric research; MAP: Military surveillance; LAP: Last-mile deliveries | Design Parameters |
| **Weight** | Nano (<200 g), Micro (<5 kg), Light (5–50 kg), Medium (50–200 kg), Heavy (>2000 kg) | Black Hornet, Avenger UAV | Micro: Portable with basic sensors; Heavy: Capable of carrying significant payloads | Nano: Surveillance; Heavy: Weapon systems, logistic transportation | Performance-Based Classification |
| **Range** | Very Short (<5 km), Short (5–50 km), Medium (50–200 km), Long (200–500 km), Ultra Long (>2000) | Tello Drone, Global Hawk | Ultra Long: Intercontinental range; Short: Indoor or local delivery | Indoor applications, delivery, border patrol, and strategic missions | Performance-Based Classification |
| **Wing Type** | Multi-Rotor, Fixed-Wing, Single Rotor | DJI Inspire, Yamaha RMAX | Multi-Rotor: High maneuverability; Fixed Wing: Aerodynamically efficient | Aerial photography, long-range mapping, agricultural spraying | Design Parameters |
| **Flying Mechanism** | Fixed Wing, Multicopter, Tilt-Wing/Tilt Rotor | SenseFly eBee, Joby S4 | Tilt-Rotor: Combines VTOL with long-range flight | Mapping, inspections, VTOL for urban environments | Design Parameters |
| **Payload** | Feather (<100 g), Light (150–270 g), Middle (400–1460 g), Heavy (>1460 g) | RoboBee, Yuneec H920 | Feather: Lightweight sensors; Heavy: Multi-sensor industrial use | Lightweight cameras, LiDAR mapping, industrial loads | Technical Attributes |
| **Application** | Military, Commercial, Industrial, Environmental, Medical, Recreational | MQ-9 Reaper, DJI Inspire 2 | Military: Target acquisition; Medical: Emergency medical supplies | Multi-sector adaptability including surveillance, delivery, and conservation | Application-Oriented Classification |
| **Power Source** | Battery, Fuel, Solar, Hydrogen Fuel Cell, Hybrid | DJI Mavic 3, Zephyr S | Battery: Lightweight; Solar: Prolonged flight; Fuel Cell: High endurance | Sustainable solutions, long-endurance flights | Performance-Based Classification |
| **Autonomy Level** | Manual, Semi-Autonomous, Fully Autonomous, Swarm | Tello Drone, Skydio 2 | Fully Autonomous: AI-driven tasks; Swarm: Multi-agent missions | Human-free operations, collaborative group missions | Engineering Specifications |
| **Communication** | LOS (Line-of-Sight), BVLOS (Beyond Visual Line-of-Sight), Satellite-Controlled, 5G-Enabled | DJI Mini SE, Global Hawk | BVLOS: Extended delivery range; Satellite: Military grade | Enhanced situational awareness, real-time communication | Operational Characteristics |
| **Sensors** | Optical, Thermal, LiDAR, Multispectral, Gas | DJI Phantom 4, FLIR drones | LiDAR: Precise topography; Thermal: Heat detection | Agricultural health monitoring, search-and-rescue missions | Technical Attributes |
| **Speed** | Low-Speed (<50 km/h), Medium-Speed (50–150 km/h), High-Speed (>150 km/h) | DJI Mini 3, Jet-powered drones | High-Speed: Tactical operations; Low-Speed: Indoor inspections | Quick responses, long-distance operations | Performance-Based Classification |
| **Durability** | All-Weather, Underwater, Ruggedized | DJI Matrice 300, BlueROV2 | All-Weather: Resists rain and high winds; Underwater: Robust water-tight designs | Extreme weather, aquatic exploration | Technical Attributes |
| **Payload Adaptability** | Fixed, Swappable | DJI Mini 2, DJI Matrice 600 | Swappable: Adaptable across multiple industries | Multi-use drones with interchangeable sensors | Design Parameters |
| **Flight Environment** | Indoor, Outdoor, Underwater, Space | Tello Drone, NASA Mars Helicopter | Underwater: Marine exploration; Space: Handles low-gravity conditions | Environment-specific operations, from indoor reconnaissance to extraterrestrial exploration | Operational Characteristics |
| **Mission Complexity** | Single Mission, Multi-Mission | DJI Mavic Mini, Matrice 300 RTK | Single: Basic imagery; Multi: Surveys, logistics, emergency response | Complex projects requiring advanced coordination | Application-Oriented Classification |

## 4. Architecture of Drone

Fig.7 presents a novel hierarchical architecture for drone systems, systematically dividing their functionalities into distinct layers. This approach separates responsibilities, enabling developers to focus on specific areas independently, thereby facilitating modular design and optimization.

At the base, the **Physical Layer** consists of core hardware like motors and sensors. The **Control Layer** manages stabilization and basic operations, while the **Communication Layer** enables data exchange between the drone and external systems. The **Navigation and Control Layer** handles flight planning and obstacle avoidance, ensuring precise maneuvering. The **Perception Layer** processes sensor data to enhance situational awareness. Above it, the **Data Link Layer** ensures secure and reliable data transmission. The **Application Layer** supports task-specific operations such as mapping or delivery. At the top, the **Security Layer** addresses cybersecurity, ensuring operational safety and data integrity. This hierarchical structure enables modular upgrades, enhances interoperability, and supports diverse applications across industries. It represents a scalable and future-ready framework for modern drone systems. operations

### 4.1 Layer Description

#### 4.1.1 Physical Layer

The Physical Layer serves as the foundation of a drone, encompassing essential components such as the frame, motors, propellers, battery, sensors, actuators, and camera systems. The frame is typically designed using lightweight and durable materials like carbon fiber to ensure both structural integrity and agility during flight. Power systems, including batteries such as Lithium Polymer (LiPo), are chosen for their high energy density and efficiency, directly influencing the drone's flight time and performance. The sensor suite, comprising gyroscopes, accelerometers, and GPS modules, provides critical data for stability, navigation, and orientation. Fig.8 shows a diagrammatic view of components. Together, these elements establish the drone's fundamental capabilities, impacting its flight dynamics, durability, and ability to carry out various operations effectively [119].

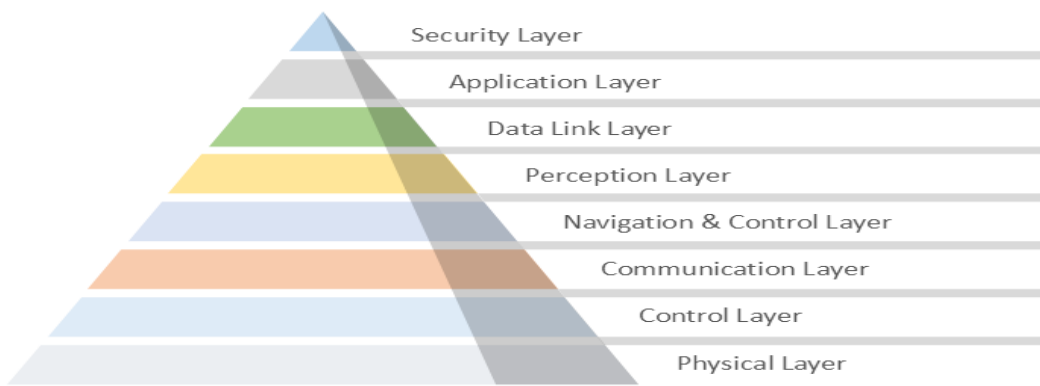

*Figure 7: Layers of Drone Architecture*

#### 4.1.2 Control Layer

The Control Layer is responsible for managing the drone's flight and stability by integrating software and hardware systems. A key component of this layer is the flight controller, which serves as the brain of the drone. Popular options include Pixhawk, DJI Naza, and ArduPilot, all of which use advanced processors like ARM Cortex and provide multiple GPIO pins for connectivity. These flight controllers support features such as PWM control for motor management, telemetry data for real-time monitoring, and custom programming for specific applications. This information is also supported by [120][121][122]

The autopilot system, powered by algorithms such as Proportional-Integral-Derivative (PID) controllers, ensures precise control of the drone's attitude, altitude, and velocity. This allows the drone to respond dynamically to environmental conditions, maintaining stability during flight [123]. Additionally, stabilization mechanisms are critical, leveraging sensors like the Inertial Measurement Unit (IMU) for real-time adjustments. Advanced filtering algorithms, such as complementary or Kalman filters, are used to minimize noise in sensor data and provide accurate readings. This information is also supported by [124]. Fig. 9 shows component of control layer.

Together, the components of the control layer ensure that the drone operates with precision, stability, and safety, dynamically adjusting its behaviour to prevent crashes and maintain smooth flight performance in varying conditions. This layer is essential for enabling the drone to execute basic manoeuvres as well as complex missions effectively This information is also supported by [125].

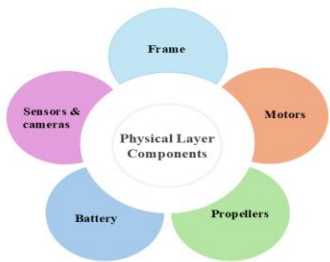

### 4.1.3 Communication Layer

The Communication Layer manages the exchange of data between the drone and external systems or operators, enabling effective remote operation and coordination. Key components include transmitters and receivers, which operate on frequency bands such as 2.4 GHz (standard) and 5.8 GHz (used for HD video transmission). Consumer drones typically have a communication range of 1–10 km, while military-grade systems can exceed 20 km. Telemetry systems, using protocols like MAVLink [126], provide real-time data transmission with data rates of up to 115200 baud for reliability. Communication protocols such as Wi-Fi, Bluetooth, 4G/5G, or proprietary RF systems ensure seamless connectivity, while encryption methods like AES-256 guarantee secure communication. This layer plays a critical role in enabling live video feeds, multi-drone coordination, and the secure exchange of mission-critical data. Fig. 9 shows component of control layer.

This is information is an inference of [127][129]

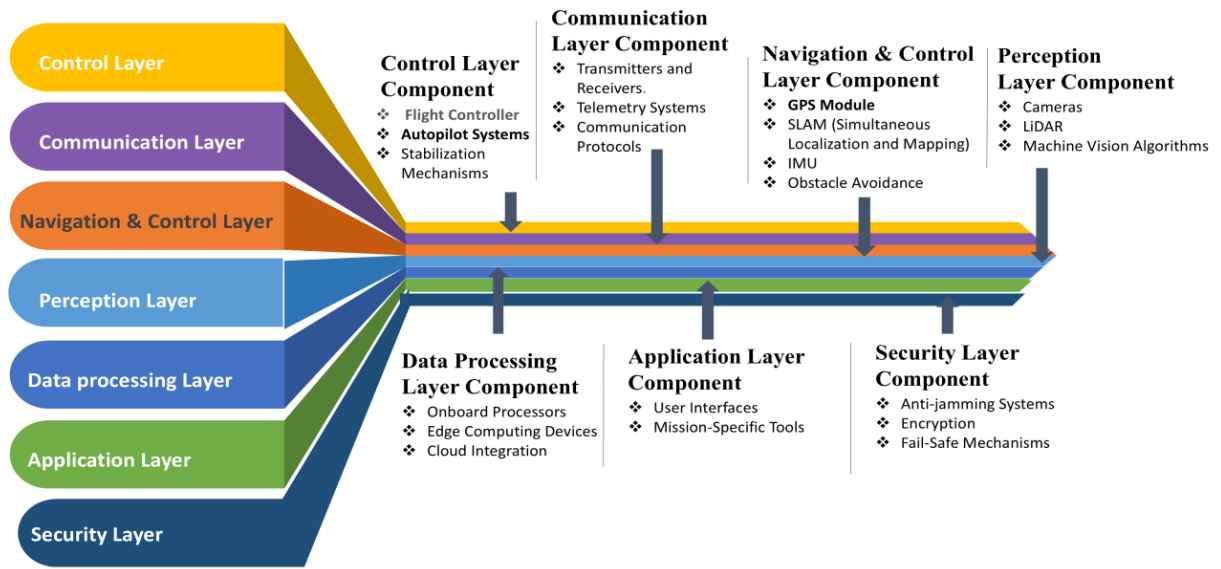

*Figure 9: Components of Layers of Drone Architecture [117...131]*

### 4.1.4 Navigation and Localization Layer

The Navigation and Localization Layer ensures the drone can determine its position and follow planned trajectories with precision. Central to this layer is the GPS module, featuring dual-band GPS (L1/L2) with an accuracy of ±1 meter and support for systems like GLONASS and Galileo for redundancy. For navigation in GPS-denied environments, SLAM (Simultaneous Localization and Mapping) is employed, utilizing sensors such as LiDAR and vision-based cameras. Algorithms like ORB-SLAM and RTAB-Map support efficient mapping and localization. Additionally, the IMU (Inertial Measurement Unit) measures linear acceleration and angular velocity with high precision (±0.02 degrees/second for advanced systems). Obstacle avoidance further enhances safety, leveraging ultrasonic, LiDAR, and infrared sensors, along with path-planning algorithms such as Rapidly exploring Random Trees (RRT) and Dijkstra's algorithm. This layer is vital for maintaining safe and reliable operations, particularly in dynamic environments or areas without GPS coverage. Fig. 9 shows component of navigation and localization layer.

Above information is concluded from [129][131]

### 4.1.5 Perception Layer

The Perception Layer allows the drone to interpret its surroundings and make autonomous decisions based on environmental data. This layer relies on various components, including cameras, which may be RGB, stereo, or thermal, offering resolutions ranging from 1080p to 4K and frame rates between 60–120 fps. LiDAR systems provide high-precision measurements with a range of 0.1–100 meters and an accuracy of ±2 cm, supporting applications like environmental mapping and 3D obstacle detection. The layer also employs machine vision algorithms, such as YOLO and Faster R-CNN, for object detection and recognition. Advanced processing units, including NVIDIA Jetson Nano and Intel Movidius, handle the computational demands of these algorithms. Together, these components enable the drone to perform tasks such as object tracking, mapping, and autonomous navigation, significantly enhancing its ability to operate effectively in complex environments. Fig. 9 shows component of perception layer. This information is an inference of [128][131].

### 4.1.6 Data Processing Layer

The Data Processing Layer is critical for processing sensor data and enabling real-time decision-making in drones. This layer incorporates onboard processors like ARM Cortex-M7 and NVIDIA Jetson series, which offer high-performance specifications such as clock speeds exceeding 1 GHz and RAM capacities ranging from 1 to 8 GB. For low-latency processing, edge computing devices play a significant role, supporting AI model deployment using frameworks like TensorFlow Lite and ONNX. Additionally, the layer integrates with cloud services through communication protocols such as MQTT and HTTP/HTTPS, leveraging platforms like AWS IoT Core and Microsoft Azure IoT Hub to facilitate seamless IoT communication. This integration allows for advanced data analytics and post-mission analysis, supporting intelligent drone operations and improving decision-making efficiency. Together, the components of this layer enable drones to process data locally and in the cloud, making them capable of handling complex tasks autonomously while ensuring efficient resource utilization. Fig. 9 shows component of Data processing Layer. This study is engraved from [129][131].

### 4.1.7 Application Layer

The Application Layer defines the operational role of a drone by providing the tools and interfaces necessary for specific missions. This layer includes user interfaces, such as mobile apps and desktop software, that offer features like real-time monitoring and route planning to enhance operational control. Fig. 9 shows component of application layer. Mission-specific tools are tailored for various applications, such as crop monitoring systems in agricultural drones, enabling precision farming. Platforms supporting these tools can be open source, like Drone-Kit, which allows for extensive customization, or proprietary solutions like DJI SDK, designed for streamlined integration with commercial systems. This layer is crucial for enabling drones to perform diverse functionalities, including delivery services, surveillance operations, and environmental monitoring, ensuring adaptability to a wide range of use cases. This layer information is supported by [129][131].

### 4.1.8 Security Layer

Fig. 9 shows component of Security Layer. The Security Layer is optional but critical for sensitive drone applications, providing essential protections against cyber-attacks and ensuring operational reliability. Key components include anti-jamming systems, which utilize methods such as frequency hopping and spread spectrum techniques to mitigate interference and maintain reliable communication. Encryption plays a crucial role in securing data, with standards like AES-256 ensuring secure communication and SSL/TLS protocols safeguarding cloud-based data exchanges. Additionally, fail-safe mechanisms such as return-to-home (RTH) and emergency landing features are incorporated to enhance safety in case of signal loss or system malfunctions. This layer is vital for protecting drone operations in security-sensitive environments, ensuring that drones can operate safely and securely in both routine and emergency. Reference of above information is taken from [130][131].

### 4.2 Summary Table of Drone Architecture

Table 14 presents a novel approach to summarizing the architecture of drone systems by explicitly defining each layer of the architecture and its corresponding components, technical specifications, and operational significance

*Table 14: Summary of Architectural layers of Drones*

| Layer | Key Components | Technical Specifications | Significance |
|---|---|---|---|
| Physical Layer | Frame, motors, propellers, battery, sensors | Lightweight (0.5–1.5 kg), 4–12 inch props, 3S–6S LiPo batteries | Determines agility, payload capacity, and durability. |
| Control Layer | Flight controller, autopilot, stabilization systems | ARM Cortex processors, PID controllers | Ensures flight stability, real-time control, and dynamic adjustments. |
| Communication Layer | Transmitters, receivers, telemetry systems | 2.4 GHz/5.8 GHz bands, range: 1–10 km | Enables remote operation, secure communication, and data transfer. |
| Navigation Layer | GPS, SLAM, IMU, obstacle avoidance | GPS accuracy ±1m, SLAM: RTAB-Map, IMU: ±0.02°/s | Provides precise positioning, navigation, and obstacle avoidance. |
| Perception Layer | Cameras, LiDAR, ultrasonic sensors | 4K cameras, LiDAR range: 0.1–100m, ±2 cm accuracy | Enhances situational awareness, enabling autonomy and decision-making. |
| Data Processing Layer | Onboard processors, edge devices, cloud integration | ARM Cortex-M7, NVIDIA Jetson, TensorFlow Lite | Facilitates AI-driven operations and real-time data analysis. |
| Application Layer | APIs, user interfaces, mission-specific tools | DJI SDK, Open-source platforms like DroneKit | Defines the operational role (e.g., surveillance, delivery). |
| **Security Layer** | Anti-jamming, encryption, fail-safe systems | AES-256 encryption, frequency hopping | Ensures operational safety, security, and cyber-attack protection. |

### 4.3 Major Drone Components of Drone

Fig.10. shows major components of Drone.

### 4.3.1 Propeller

The propellers are vital for generating thrust and enabling the drone to achieve lift, stabilize itself, and maneuver in flight. Made of lightweight materials such as carbon fiber or plastic, they come in various sizes, typically ranging from 4 to 12 inches. The propeller design directly influences the drone's aerodynamic efficiency and power consumption, impacting its flight dynamics. Fig. 11 shows the type of propellers used in drone. Table 15 gives the summary of types of propellers with material , specifications and applications in breif. Above information is concluded from [132][133][134][139][140] [141][142][143][144]

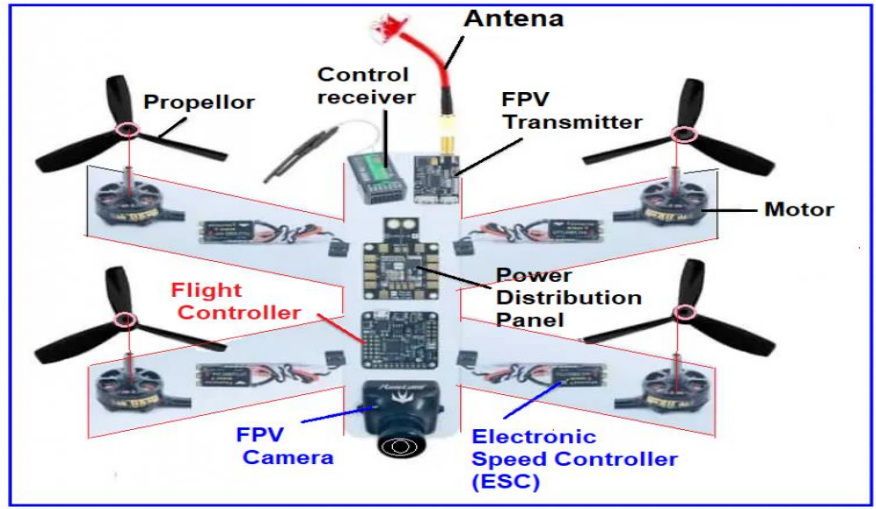

*Figure 10: Major Components of Drone. [132]*

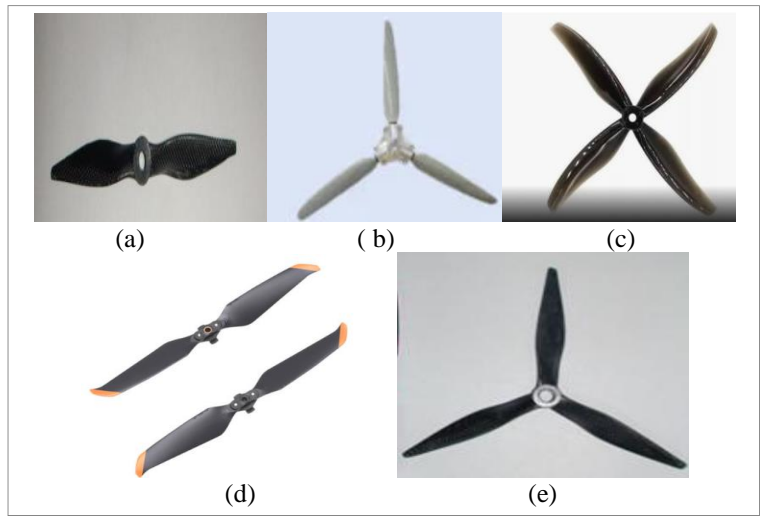

*Figure 11:Propeller Types (a) Two-Blade [135] (b) Variable-Pitch [133]. (c) Four- Blade [138]. (d)Low-Noise [137]. (e) Three Blade [136]*

*Table15: Summary Table of Types of Propellers [139][140][141][142][143][144]*

| Propeller Type | Material | Technical Specifications | Applications |
|---|---|---|---|
| **Two-Blade Propeller** | Plastic, Carbon Fiber | Lightweight; High agility; Moderate thrust | Racing drones, hobby UAVs |
| **Three-Blade Propeller** | Plastic, Aluminum Alloy | Higher thrust; Reduced efficiency; Moderate noise | Photography drones, delivery drones |
| **Four-Blade Propeller** | Reinforced Plastic, Carbon Fiber | Enhanced thrust; High power requirement; Increased stability | Industrial drones, precision agriculture |
| **Low-Noise Propeller** | Composite | Noise reduction features; | Surveillance, consumer |

| | Materials, Polymer | High durability | drones |
|---|---|---|---|
| **Variable-Pitch Propeller** | Carbon Fiber, Aluminum Alloy | Adjustable blade angle; Optimized for efficiency and power | Military-grade UAVs, heavy-lift drones |

### 4.3.2 Motor

Fig. shows type of motors and Table 16. Gives motors description with applications. Motors drive the propellers by converting electrical energy into mechanical energy, which is essential for lift and maneuvering. Most drones use brushless DC motors due to their high efficiency, longevity, and reduced maintenance requirements. These motors are designed to offer precise speed control and are critical for achieving flight stability. Motors enables drones to maintain accurate control during dynamic maneuvers and their impact on flight endurance. Motor as a component of drone is referenced from [132][139][140][145][146][147].

### 4.3.3 Electronic Speed Controller (ESC)

The ESC regulates the motor's speed, providing the necessary control for stable flight. It interprets signals from the flight controller and adjusts motor performance accordingly. ESCs support PWM (Pulse Width Modulation) signals and are tailored to match the drone's motor and power system specifications. Efficient ESCs are critical for synchronizing motor speed and ensuring that the drone responds accurately to control inputs. ESC as drone component were concluded from [132][139][140]

*Table 16: Summary of Types of Motors [139][140][145][146][147]*

| Motor Type | Description | Specifications | Applications |
|---|---|---|---|
| **Brushed Motors** | Simple DC motors with brushes for current transfer. | Voltage: 3-12V; Efficiency: ~50%; Lifetime: ~1000 hours | Toy drones, small indoor drones |
| **Brushless DC Motors (BLDC)** | Motors without brushes; rely on electronic commutation using controllers. | Voltage: 7.4V-24V; KV rating: 500-3000 KV; Efficiency: ~85% | Consumer drones, racing drones, photography drones |
| **Hybrid Motors** | Combine electric motors with combustion engines for extended flight duration and higher payloads. | Combustion engine capacity: 10-30cc; Electric motor power: 50-200W | Long-endurance drones, industrial UAVs |
| **Switched Reluctance Motors (SRM)** | Electrically commutated motors with simple, robust designs and low maintenance. | Power: 50-500W; Torque: Moderate; Efficiency: ~75% | Experimental drones, cost-effective UAVs |

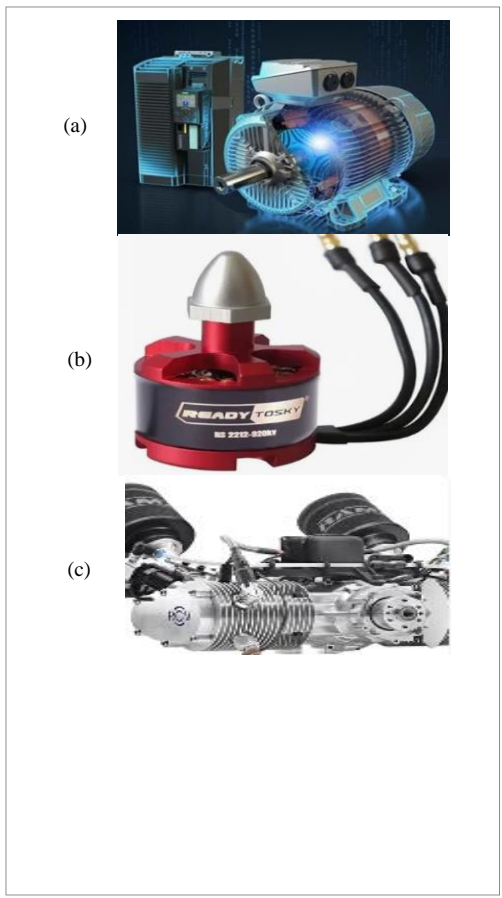

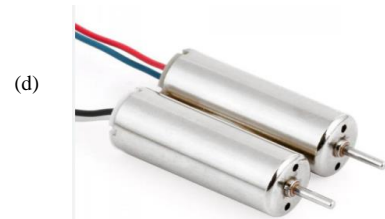

*Figure 12:Motor Types (a)SRM [148] (b) BLDC [149] (c)Hybrid [150]. (d)Brushed [151]*

### 4.3.4 Flight Controller

Fig. 12 shows the images of FC discussed in summary Table 17. Acting as the drone's brain, the flight controller processes sensor inputs from devices such as gyroscopes and accelerometers. It dynamically adjusts motor speeds to maintain stability and execute manoeuvring commands. Equipped with IMUs and GPS modules, it ensures precise navigation and stabilization. Flight controllers act as the backbone of drone automation, enabling autonomous flight and enhanced stability even in adverse conditions. FC as drone component were concluded from [132][139][140][157][158].

### 4.3.5 FPV (First-Person View) Camera

The FPV camera provides real-time video feeds to operators, facilitating navigation, surveillance, and other visual tasks. With resolutions ranging from 1080p to 4K and frame rates of 30–120 fps, FPV cameras offer clear and detailed visuals. FPV cameras enhance situational awareness, making them indispensable for tasks like search-and-rescue missions or real-time reconnaissance. FPV as drone component were concluded from [133][139][140].

### 4.3.6 Control Receiver (CR)

The control receiver establishes communication between the drone and the operator's remote controller. It processes signals sent by the controller and relays them to the flight controller for execution. Operating on frequency bands such as 2.4 GHz, the receiver ensures reliable and interference-free control. CR as drone component were concluded from [139][140][163]

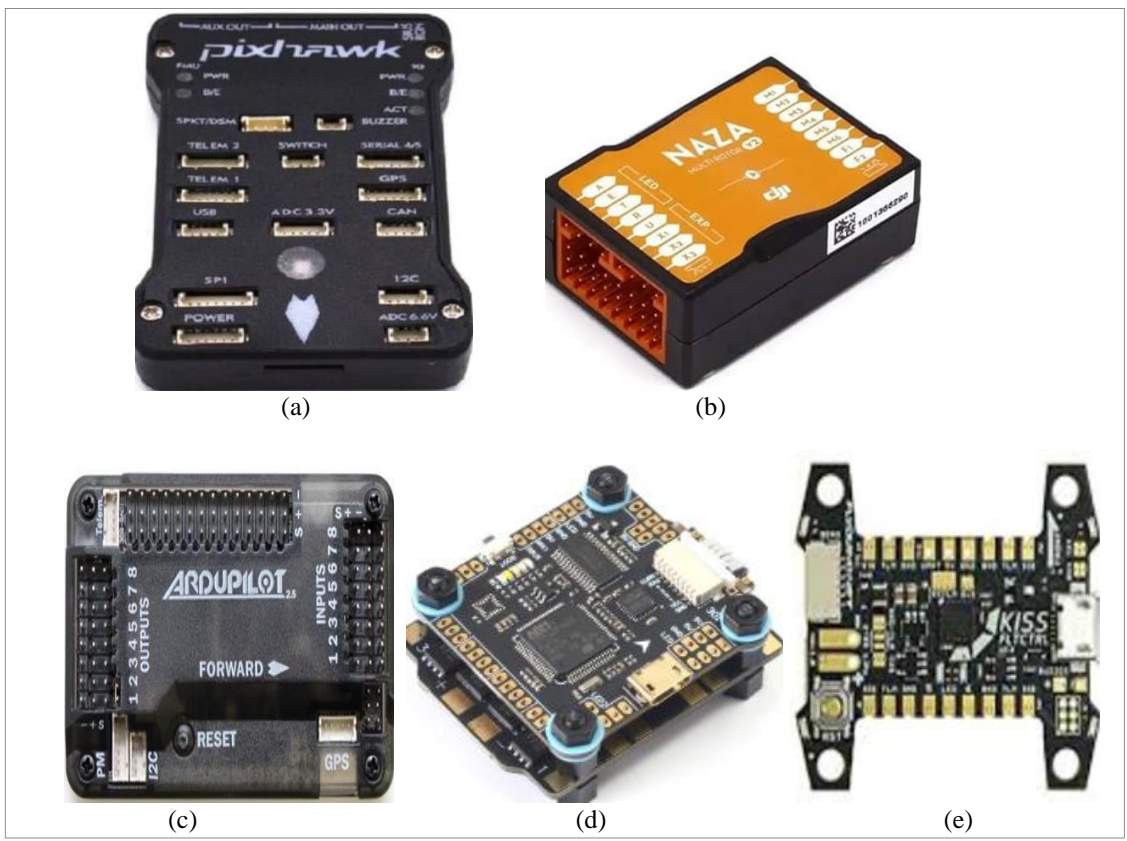

*Figure 13: FC Types (a) Pixhawk [152] (b) DJI Naza [153] (c) ArduPilot [154] (d) BetaFlight [155] (e) KISS[156]*

*Table 17: Summary of Types of Flight Controllers [132][139][140][157][158]*

| Flight Controller Type | Description | Specifications | Applications |
|---|---|---|---|
| **Pixhawk** | Open-source flight controller supporting multiple platforms like Ardu Pilot and PX4. | Processor: STM32F427 Cortex-M4, 256 KB RAM; Sensors: IMU, barometer | Professional drones, custom drone projects |
| **DJI Naza** | Proprietary flight controller with integrated GPS and stabilization features. | GPS-enabled, supports multi-rotor configurations; Weight: ~25g | Consumer drones, FPV drones |
| **Ardu Pilot** | Open-source controller supporting diverse platforms (fixed-wing, multirotor, etc.). | Processor: ARM Cortex-M3; Interfaces: I2C, UART, SPI | Research applications, autonomous drones |
| **Beta Flight** | Flight controller firmware optimized for racing drones and FPV performance. | High refresh rate; supports DSHOT ESC protocol | Racing drones, freestyle FPV drones |
| **KISS (Keep It Simple Stupid)** | Minimalistic controller designed for performance and ease of use. | ESC Protocol: KISS ESC; Low latency; Lightweight (~5g) | Racing drones, beginner drone pilots |
| **INAV** | Advanced flight controller firmware supporting navigation and waypoint missions. | Processor: STM32F405; Features: GPS integration | Long-range drones, navigation drones |
| **APM (Autopilot Module)** | Legacy open-source controller with support for advanced autopilot features. | Processor: Atmel AVR; Interfaces: PWM, I2C, UART | Agricultural drones, mapping drones |

### 4.3.7 FPV Transmitter

The FPV transmitter enables the live video feed captured by the camera to be transmitted to the operator in real-time. Using frequency bands such as 5.8 GHz, it provides a low-latency and high-quality transmission channel. FPV transmitters are vital for applications requiring real-time situational awareness, such as drone racing or military surveillance operations FPV transmitter as drone component were concluded from [139][140].

### 4.3.8 Power Distribution Panel

The power distribution panel ensures that electrical power from the drone's battery is distributed evenly across all components, including the motors, ESCs, and flight controller. Designed to handle high current loads, it stabilizes voltage output, ensuring the safe and efficient operation of onboard systems. The importance of reliable power distribution systems in maximizing drone performance and operational safety. Above information about power distribution is supported by [139][140].

### 4.3.9 Antenna

The antenna enhances the quality and range of communication between the drone and external systems. Designed for specific frequency bands, such as 2.4 GHz or 5.8 GHz, antennas help minimize interference while extending the operational range. Advanced antenna systems are essential for ensuring reliable data transmission during long-range missions or in challenging environments. Table 18 gives the summary of types of Antenna and their description. Antenna as drone component were concluded from [139][140][159].

*Table 18: Summary of Types of Antennas [159][160][161][162]*

| Antenna Type | Description | Specifications | Applications |
|---|---|---|---|
| Omnidirectional Antenna | Radiates signals uniformly in all directions, ensuring consistent connectivity. | Frequency: 2.4 GHz/5.8 GHz; Gain: 2–5 dBi | Consumer drones, FPV racing drones |
| Directional Antenna | Focuses signal in a specific direction, enhancing range and signal strength. | Frequency: 5.8 GHz; Gain: 8–15 dBi | Long-range drones, surveillance drones |
| Patch Antenna | Compact, flat antennas ideal for GPS and precise communication. | Frequency: 1.575 GHz (GPS L1), Gain: ~3 dBi | GPS modules in drones |
| Helical Antenna | Spiraled antennas providing circular polarization for robust signal in diverse orientations. | Frequency: 2.4 GHz; Gain: 10–15 dBi | FPV drones, high-altitude drones |
| Yagi-Uda Antenna | Directional antenna with high gain, typically used for extended range. | Frequency: 5.8 GHz; Gain: 10–18 dBi | Long-range communication, telemetry |
| Dipole Antenna | Simple, robust design used for basic RF communication. | Frequency: 2.4 GHz/5.8 GHz; Gain: 2–5 dBi | Consumer drones, indoor drones |
| Parabolic Antenna | Parabolic reflectors offering high directional gain for point-to-point communication. | Frequency: 2.4 GHz; Gain: 15–25 dBi | Surveillance drones, military applications |
| MIMO Antenna | Multiple-input, multiple-output antennas for improved data throughput and range. | Frequency: 2.4 GHz/5 GHz; Multiple antenna elements | High-speed drones, video transmission |

## 5. NAVIGATION AND CONTROL SYSTEM OF DRONE

### 5.1 Drone Navigation System

Drone navigation system integrates advanced technologies to enable precise movement, path planning, and mission execution in various environments. It combines hardware (e.g., sensors and actuators) with sophisticated software algorithms for autonomy and reliability. Fig.14 gives major drone Navigation components. Below is a breakdown of critical components and functionalities.

### 5.1.1 GPS-Based Navigation

Drones leverage Global Positioning Systems (GPS) for geolocation and adherence to predefined routes, with dual-frequency GPS modules (L1/L2) enhancing accuracy and reducing signal interference. These capabilities make drones highly effective in outdoor applications such as surveying, delivery services, and search-and-rescue missions. However, their functionality is limited in urban or indoor environments where GPS signals are prone to interference. This information is supported by the studies of [164]

### 5.1.2. Inertial Navigation System (INS)

Inertial Measurement Units (IMUs), which integrate accelerometers and gyroscopes, are essential components in modern drones. They measure motion parameters such as velocity and angular rates, enabling dead reckoning when GPS signals are unavailable. This functionality is particularly useful for enhancing drone performance in GPS-denied environments, including tunnels or dense forests. These insights are supported by the studies of [165][168][179].

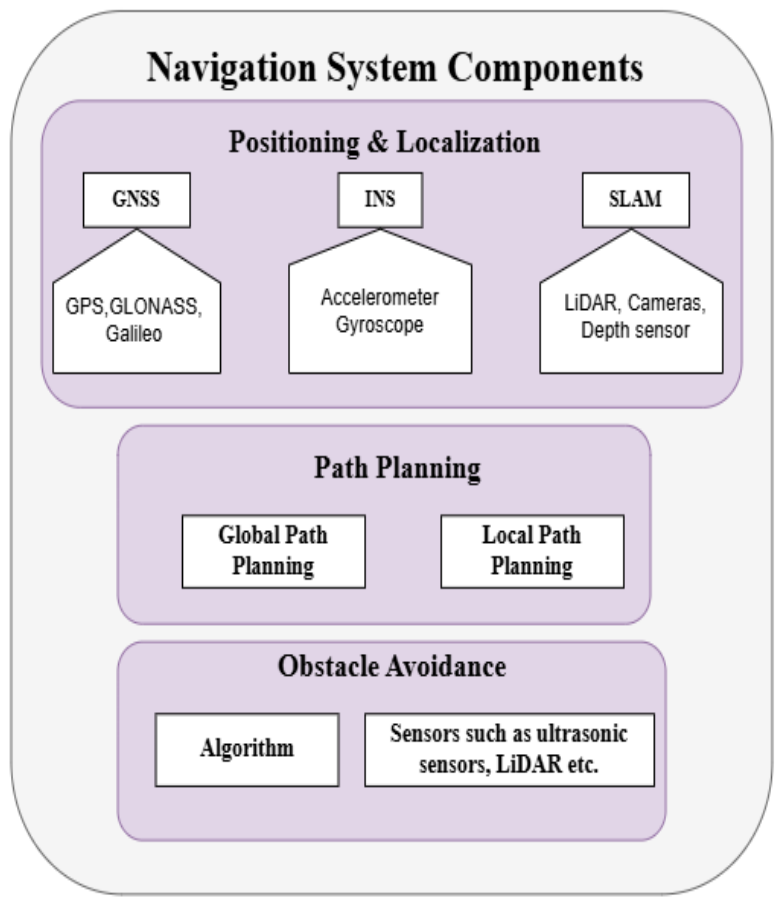

Figure 14: Navigation System Components [165][166][168][171][179]

### 5.1.3. Simultaneous Localization and Mapping (SLAM)

Drones use simultaneous localization and mapping (SLAM) techniques to combine sensor data, such as LiDAR and vision cameras, for creating an environmental map while simultaneously determining their position within it. Common algorithms employed for this purpose include ORB-SLAM and RTAB-Map. These capabilities are crucial for indoor navigation, autonomous inspections, and obstacle avoidance. This information is supported by studies from [165][168].

### 5.1.4 Path Planning Algorithms

Path planning for drones is categorized into global and local path planning. Global path planning determines optimal routes using algorithms such as A* and Dijkstra, while local path planning focuses on real-time adjustments using methods like Rapidly exploring Random Trees (RRT) to avoid obstacles. Together, these approaches ensure efficient and safe navigation by dynamically recalibrating routes as needed. This classification and functionality are discussed in [165][166][170][174]

### 5.1.5. Obstacle Detection and Avoidance

Drones utilize path planning techniques that are broadly divided into global and local approaches. Global path planning employs algorithms like A* and Dijkstra to determine optimal routes, while local path planning relies on methods such as Rapidly exploring Random Trees (RRT) to make real-time adjustments and avoid obstacles. These complementary strategies work together to ensure efficient and safe navigation by dynamically recalibrating routes as conditions change. The principles and effectiveness of these methods are explored in the studies [166]

### 5.1.6 Machine Learning and AI Integration

Artificial intelligence (AI) models play a crucial role in drone navigation by analyzing sensor data to predict environmental changes and optimize navigation strategies. This capability is particularly valuable in complex scenarios, such as rescue operations or coordinating swarming behavior in multi-drone systems. The applications and advancements in this area are highlighted in the studies by [180]

### 5.1.7 Edge-Assisted Navigation

Edge computing enhances drone functionality by offloading computational tasks from onboard processors to edge servers, significantly improving data processing speed and reducing latency. This approach is particularly effective for high-speed drones or missions that demand real-time analytics. The advancements in this technology are detailed in the study [166].

### 5.1.8 Visual Navigation

Vision-based navigation in drones relies on components such as stereo cameras and optical flow sensors. These systems use image processing to detect environmental features and determine relative motion, making them particularly suitable for GPS-denied zones like indoor arenas or confined spaces. The effectiveness of this approach is supported by studies from [166].

### 5.2. Control System

The control system of a drone is a critical component that ensures stable flight, accurate navigation, and effective communication. Ref. Fig 15 gives major component of control system. It comprises three main subsystems: the Flight Controller, the Stabilization Mechanism, and Communication Systems [179].

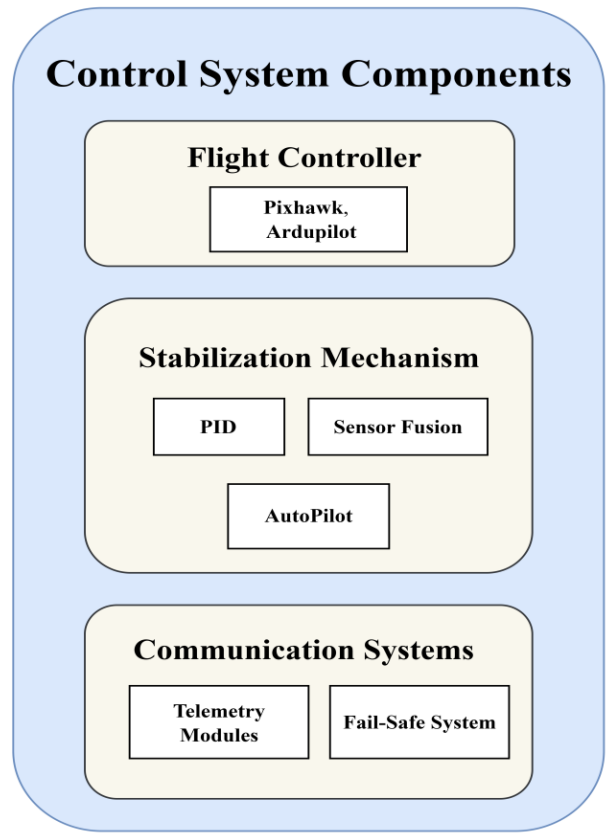

*Figure 15: Control System Components [179][181]*

### 5.2.1. Flight Controller
The Flight Controller, which acts as the central processing unit (CPU) of the drone, is responsible for managing inputs from various sensors and executing precise flight commands. Popular open-source platforms like Pixhawk and ArduPilot are widely utilized due to their advanced features, including waypoint navigation, obstacle avoidance, and real-time telemetry data logging. These controllers integrate seamlessly with other drone systems, enabling efficient operation across diverse applications. This information is an inference of [181]

### 5.2.2 Stabilization Mechanism
The Stabilization Mechanism is essential for maintaining balance and smooth operation, even in challenging conditions. It employs PID (Proportional-Integral-Derivative) controllers, which continuously adjust motor speeds to correct flight deviations and ensure stability. Sensor Fusion plays a pivotal role by integrating data from multiple sources, such as accelerometers, gyroscopes, and magnetometers, to provide accurate positional information. Additionally, Auto-Pilot systems enable autonomous flight by executing pre-programmed missions like waypoint navigation or terrain following without requiring manual intervention. Recent advancements in stabilization mechanisms, including AI- based sensor fusion, have enhanced drone performance in dynamic environments. These insights are supported by the studies of [181]

### 5.2.3. Communication System
The Communication Systems facilitate real-time interaction between the drone and the ground control station (GCS). Telemetry modules transmit critical data, such as position, altitude, and mission status, using protocols like MAV Link over radio frequencies or 4G/5G networks. A robust Fail-Safe System is also integrated to execute predefined safety measures, such as returning to the home location or safe landing, in cases of signal loss, power depletion, or other failures. Redundant and secure communication systems are particularly critical in high-risk or autonomous operations to prevent hacking or interference. These insights are supported by the studies of [181].

To optimize drone performance, the control system should include efficient power management, environmental adaptability, and compliance with safety regulations like those from the Federal Aviation Administration (FAA) or European Union Aviation Safety Agency (EASA). Incorporating advanced AI and machine learning algorithms for adaptive stabilization and real-time decision-making can further enhance the system's reliability and operational efficiency.

These integrated control system components are critical for the success of modern drones in applications ranging from autonomous inspections to swarm-based missions. These insights are supported by the studies of [181]

### 5.4. Drone Navigation System Categorization

Fig. 16 gives the major classification of Drone navigation systems.

### 5.4.1 Indoor Navigation Systems

Indoor navigation systems are tailored for environments where GPS signals are unavailable, such as warehouses, tunnels, or manufacturing facilities. These systems rely heavily on advanced technologies like Simultaneous Localization and Mapping (SLAM), which employs vision sensors (monocular or stereo cameras) and inertial measurement units (IMUs) to generate real-time maps while simultaneously localizing the drone. Algorithms like ORB-SLAM and RTAB-Map are commonly used for these purposes. Additionally, optical flow sensors play a significant role in detecting relative motion by analyzing shifts in consecutive images, enabling precise navigation in cluttered Indoor spaces. Fig. 17 gives the major inside components of indoor navigation system.

For path planning, drones utilize algorithms like Rapidly exploring Random Trees (RRT), which dynamically calculate paths to avoid obstacles in real time. These systems are crucial for scenarios like inventory management, surveillance, and medical supply delivery within buildings. However, they face challenges such as dynamic environments where frequent map updates are required. Supported by the studies of [165][169][170][171][172][173][178][180]

### 5.4.2 Outdoor Navigation Systems.

Outdoor navigation systems primarily depend on Global Navigation Satellite Systems (GNSS) such as GPS, GLONASS, or Galileo for real-time geolocation. These systems are effective in open environments but struggle in urban canyons or dense forests where signals are degraded [170][177].

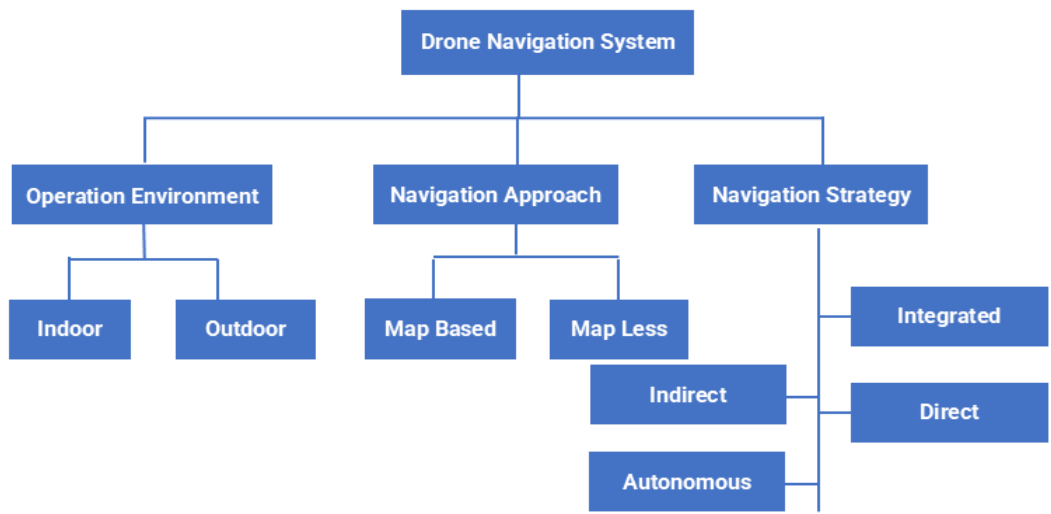

*Figure 16: Drone Navigation System Classification [165][173][178]*

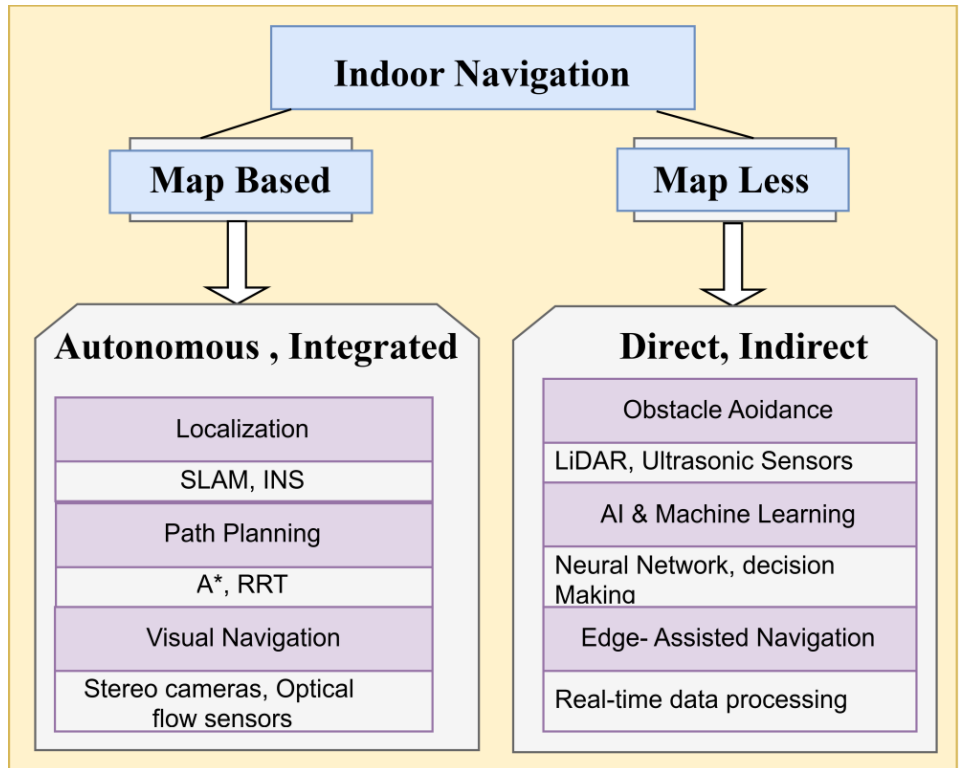

Figure 17: Detailed Classification of Indoor Navigation System [165][166][169][167][170][172][173][178]

To counteract this limitation, Inertial Navigation Systems (INS) are integrated with GPS to provide continuous position tracking during GPS outages. Vision-based augmentation, such as Visual SLAM, further enhances localization by providing corrections to GPS data. Path planning in outdoor navigation typically involves global algorithms like A* and Dijkstra for route optimization, ensuring efficient coverage of large areas. Additionally, obstacle avoidance systems, often powered by LiDAR or ultrasonic sensors, ensure safe navigation in dynamic outdoor environments. Outdoor navigation is widely applied in agriculture, search-and-rescue missions, and package delivery. Fig. 18 gives the major inside components of outdoor navigation system These insights are supported by the studies of [171]

### 5.4.3 Map-Based Navigation

Map-based navigation systems utilize either pre-defined maps or real-time mapping techniques. LiDAR-SLAM is a prominent method that generates highly accurate 3D point clouds, while visual SLAM offers a lightweight alternative by using camera feeds to build 2D or 3D maps. These methods are effective in stable environments but may struggle in highly dynamic conditions where frequent map updates are required. Map-based navigation is extensively used in tasks like autonomous inspections, where precision and reliability are critical. Above approach were supported by studies of [173][178]

### 5.4 4 Mapless Navigation

Mapless navigation systems rely on real-time sensor feedback rather than pre-existing maps, making them ideal for dynamic and unexplored environments. These systems use data from cameras, LiDAR, and other sensors to identify obstacles and adjust paths dynamically. AI-driven methods, such as response, where environments are unpredictable. Above conclusion were supported by studies of [174][175][176][178].

### 5.4.5 Navigation Strategies

Integrated navigation systems combine multiple navigation techniques, such as GPS, INS, and SLAM, to ensure robust and reliable performance. For example, GPS provides global positioning, INS smooths out interruptions in GPS data, and vision sensors enhance localization accuracy in diverse environments. Direct navigation systems rely on real-time sensor inputs for immediate path adjustments, making them highly effective for obstacle avoidance in dynamic scenarios. Conversely, indirect navigation systems process sensor data to improve overall navigation efficiency, often delegating control to centralized units. Lastly, autonomous navigation systems employ AI to analyze sensor data, predict environmental changes, and make real-time decisions. Table 19 gives the summary of navigation strategies

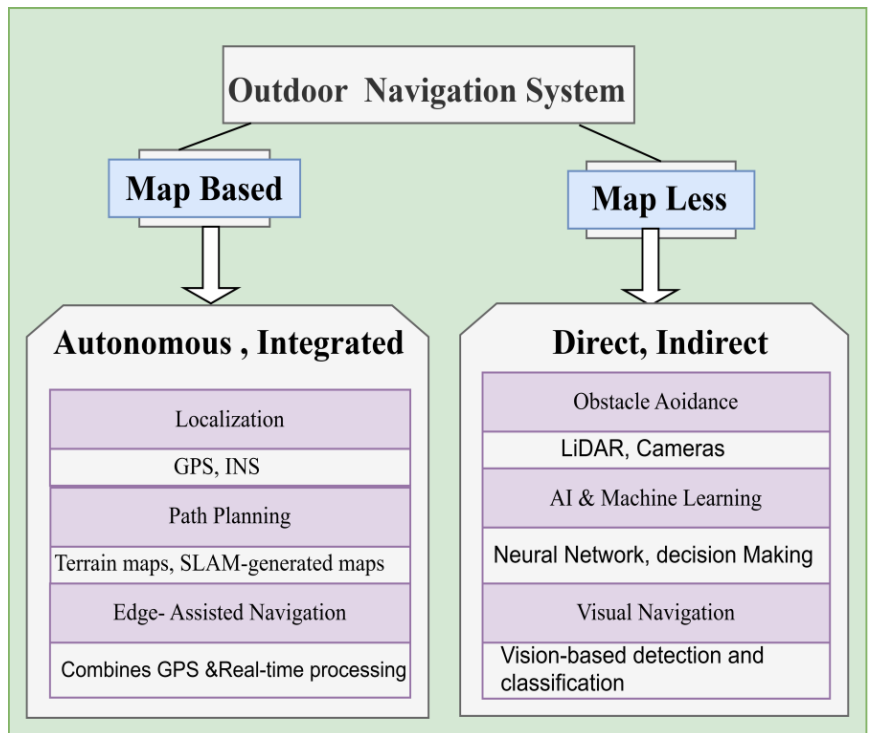

*Figure 18: Detailed Classification of Outdoor Navigation System [167] [170][171]*

These systems are used in tasks requiring high autonomy, such as multi-drone coordination and exploration missions. The above stated information were concluded from [167]

*5.4.6 Vision-Based Navigation*

Vision-based navigation is a versatile technique that integrates seamlessly across all navigation systems. Indoors, it aids SLAM/INS and obstacle avoidance, while outdoors, it complements GPS by enhancing localization accuracy. In map-based systems, vision data is used to build detailed maps, whereas in map-less systems, it facilitates dynamic obstacle detection and motion estimation. Vision-based navigation also underpins autonomous and integrated navigation systems, providing the visual data necessary for advanced AI algorithms. This study is supported by [170][172]

*Table 19: Summary of Drones Navigation System[170]...[178]*

| Navigation System | Approach | Strategies | Technologies Used |
|---|---|---|---|
| **Indoor Navigation** | Map-Based | Autonomous, Integrated | - Positioning & Localization: SLAM, INS<br>-Path Planning: A*, RRT<br>-Visual Navigation: Stereo cameras, optical flow sensors |
| | Map-less | Direct, Indirect | -Obstacle Avoidance: LiDAR, Ultrasonic sensors<br>- AI & Machine Learning: Neural networks for decision-making<br>- Edge-Assisted Navigation: Real-time data processing and computation offloading |
| **Outdoor Navigation** | Map-Based | Autonomous, Integrated | -Positioning & Localization: GPS, INS<br>-Path Planning: Preloaded terrain maps, SLAM-generated maps<br>-Edge-Assisted Navigation: Combines GPS and real-time sensor data processing |
| | Map-less | Direct, Indirect | -Obstacle Avoidance: Vision-based obstacle detection using LiDAR or cameras<br>-AI & Machine Learning: AI-driven navigation algorithms for decision-making<br>-Visual Navigation: Cameras for object detection and classification |

## 6. APPLICATION OF DRONE

Fig.19. shows the major categorization of drone Application. Details of each category is discussed in below section.

### 6.1. Agriculture

Drones have revolutionized agriculture by enabling precision farming and efficient resource management. They are used for crop health monitoring, where multispectral imaging and thermal cameras detect stress levels in plants, such as nutrient deficiencies, disease, or water stress. Drones are also pivotal in pest control, deploying automated spraying systems that apply pesticides or fertilizers precisely where needed, reducing chemical usage and environmental impact. In irrigation management, drones equipped with sensors map soil-moisture variability, optimizing water usage across fields. These advancements collectively enhance productivity and sustainability in modern agriculture. Several studies underscore the transformative role of drones in agriculture. Fig. 20 (e) gives agriculture-based sub classification of application.

[183] highlights the integration of drones in precision farming for accurate data acquisition and decision-making. [184][185] Drones also facilitate real-time monitoring of large-scale operations, enabling rapid response to issues.

### 6.2. Logistics and Delivery

Drones have become instrumental in streamlining logistics and delivery systems, especially for last-mile urban delivery, medical supply transport, and package delivery in hard-to-reach areas. They utilize GPS-based routing for accurate navigation, obstacle avoidance systems to ensure safety during transit, and AI algorithms to optimize delivery routes dynamically. For instance, autonomous drones are increasingly employed in emergency medical deliveries, drastically reducing response times in critical scenarios. About Logistic and delivery as an application of drone were discussed in [186][187].

### 6.3. Infrastructure and Construction

In construction, drones are revolutionizing building inspections, site surveys, and 3D modelling, allowing for enhanced accuracy and efficiency. Technologies like LiDAR and photogrammetry capture high-resolution data, while edge computing processes this information in real-time, enabling actionable insights. These tools are invaluable in monitoring large-scale projects and ensuring compliance with safety standards. Role of in Infrastructure and construction were studied from [188][189][190]. Fig. 20 (a) gives sub classification of application.

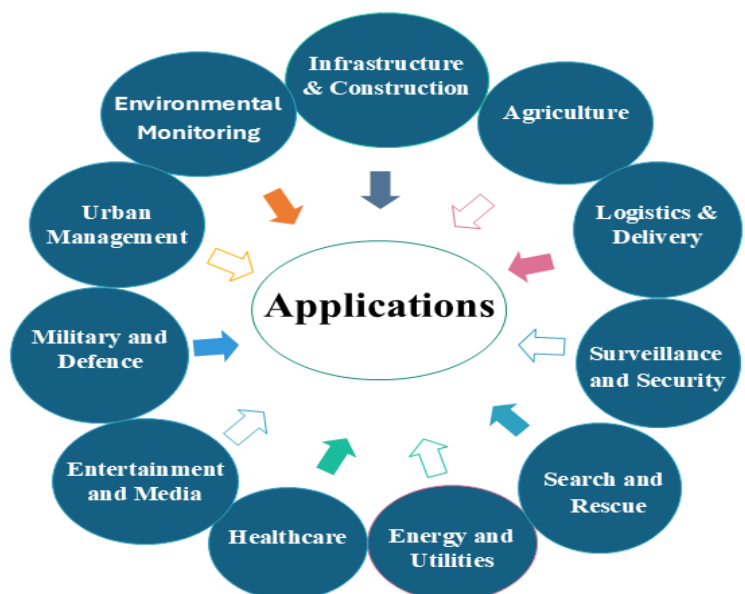

*Figure 19: Drone Application Categorization [99]*

### 6.4. Environmental Monitoring

Environmental applications of drones range from wildlife tracking and deforestation surveillance to air quality monitoring. Equipped with thermal imaging, multispectral sensors, and AI-powered analytics, drones can collect and process data for conservation efforts, ecosystem studies, and pollution tracking. Their versatility makes them critical in addressing environmental challenges effectively. Role of Drone in Environmental Monitoring was discussed in [191][192]. Fig. 20 (b) gives sub classification of application.

### 6.5. Surveillance and Security

Drones play a vital role in border surveillance, disaster response, and event monitoring. Using night vision,

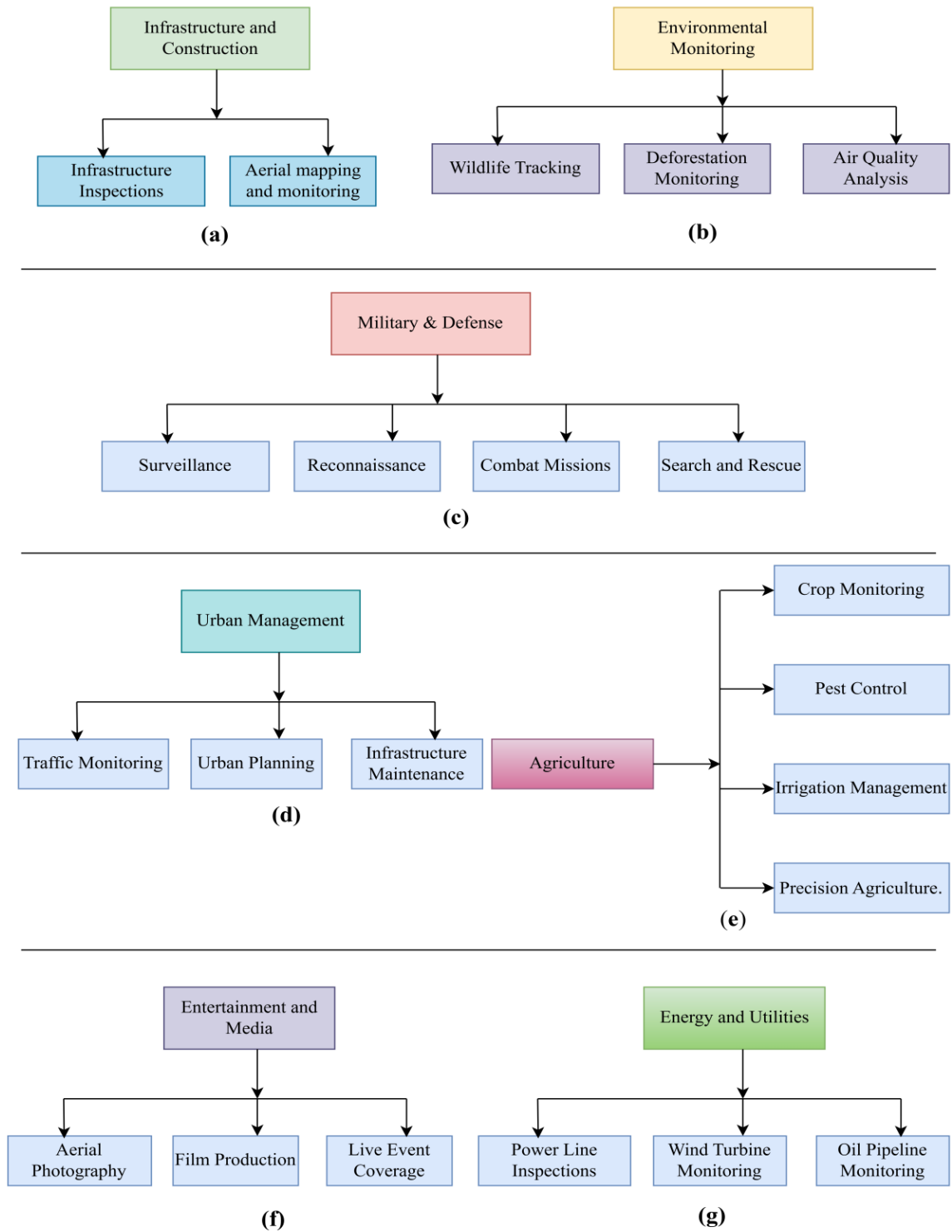

*Figure 20: Sub-categorial classification of Drone Application (a) Infrastructure and Construction (b) Environmental Monitoring (c) Military and Defence (d) Urban management (e) Agriculture based Application classification (f) Entertainment and Media (g) Energy and Utilities*

real-time streaming, and anomaly detection, they enhance situational awareness in high-risk areas. These systems provide law enforcement and security agencies with effective tools for maintaining safety and responding to emergencies. Drone in surveillance was discussed in [193][194]

### 6.6. Healthcare

In healthcare, drones support emergency medical deliveries, public health monitoring, and organ transport. Integrated with IoT, GPS navigation, and real-time analytics, these systems can deliver life-saving supplies to remote areas and improve the efficiency of healthcare operations [195][196][197][198]

### 6.7. Entertainment and Media

The media industry leverages drones for aerial cinematography, event broadcasting, and immersive experiences. Advanced technologies like high-resolution cameras, stabilization systems, and AI-based scene optimization allow filmmakers and broadcasters to capture unique perspectives, enriching the audience experience [199][200]. Fig. 20 (f) gives sub classification of application.

### 6.8. Search and Rescue

Drones are invaluable in disaster site analysis, locating missing persons, and supply delivery during crises [210][202][203].

### 6.9 Summary Table of Drone Application

Table 20. illustrate the summary of existing drone application with their real-life examples and Technology used.

*Table 20: Summary of Drone Application [226] ... [244]*

| S. No | Category | Applications | Technologies Used | Real-Life Example |
|---|---|---|---|---|
| 1 | Agriculture | Crop health monitoring, pest control, irrigation management, precision agriculture | Multispectral imaging, automated spraying, AI-based analysis | DJI Agras drones used by farmers for spraying pesticides and monitoring crop health in Europe and Asia. |
| 2 | Logistics and Delivery | Package delivery, medical supply transport, last-mile delivery | GPS-based routing, obstacle avoidance, AI optimization | Amazon Prime Air delivering packages to customers' doorsteps in the U.S. |
| 3 | Infrastructure and Construction | Building inspections, site surveys, 3D modeling | LiDAR, photogrammetry, edge computing | DJI Matrice 300 drones used in New York for inspecting high-rise buildings for maintenance and safety. |
| 4 | Environmental Monitoring | Wildlife tracking, deforestation surveillance, air quality monitoring | Thermal imaging, multispectral sensors, real-time AI analytics | Conservation drones used in the Amazon rainforest for detecting illegal logging and monitoring biodiversity. |
| 5 | Surveillance and Security | Border surveillance, disaster response, event monitoring | Night vision, real-time streaming, anomaly detection | Border patrol drones used by U.S. Customs and Border Protection for monitoring borders. |
| 6 | Healthcare | Emergency medical deliveries, public health monitoring, organ transport | IoT, GPS navigation, real-time analytics | Zipline drones delivering blood and medical supplies to remote villages in Rwanda. |
| 7 | Entertainment and Media | Aerial cinematography, event broadcasting, immersive experiences | High-resolution cameras, stabilization systems, AI scene optimization | Drones used in filming the Netflix series Our Planet for breathtaking wildlife shots. |
| 8 | Search and Rescue | Disaster site analysis, locating missing persons, supply delivery | Thermal imaging, SLAM, AI-powered object recognition | Drones deployed in Turkey for search and rescue operations after the 2023 earthquake. |
| 9 | Military and Defense | Surveillance, reconnaissance, combat missions, supply logistics | Autonomous navigation, secure communication, weaponized systems | MQ-9 Reaper drones used by the U.S. military for reconnaissance and targeted operations. |
| 10 | Urban Management | Traffic monitoring, urban planning, infrastructure inspection | AI traffic analysis, IoT, sensor-based inspections | Drones monitoring traffic congestion in Dubai for real-time urban planning updates. |
| 11 | Energy and Utilities | Monitoring pipelines, inspecting wind turbines, solar panel maintenance | Infrared imaging, LiDAR, real-time reporting | Energy companies using drones in Texas to inspect wind turbines and solar farms for efficiency. |

## 7. Key Challenges in Drone Technology

Fig. 21 categorizes the primary challenges associated with drone technology into several interconnected domains, highlighting the multifaceted nature of these issues. It serves as a visual representation to provide a structured understanding of the key hurdles that hinder the adoption and advancement of drones. The challenges are divided into the following major categories. Table 21 gives the summary of major classification of drone challenges. Below section explain each challenge in detail.

### 7.1 Technical Challenges

Technical challenges are a significant barrier to the advancement of drone technology. One of the foremost issues is **battery life and energy efficiency [205]**, as limited flight duration due to insufficient battery capacity restricts operational capabilities, necessitating innovations in energy storage for extended use. Another critical limitation is **payload capacity**, which confines drones' ability to carry heavy or multiple items simultaneously, reducing their utility in various applications. Furthermore, while progress has been made in **autonomy and AI**, achieving fully autonomous operations with reliable AI-based navigation and decision-making systems remains an ongoing challenge [206].

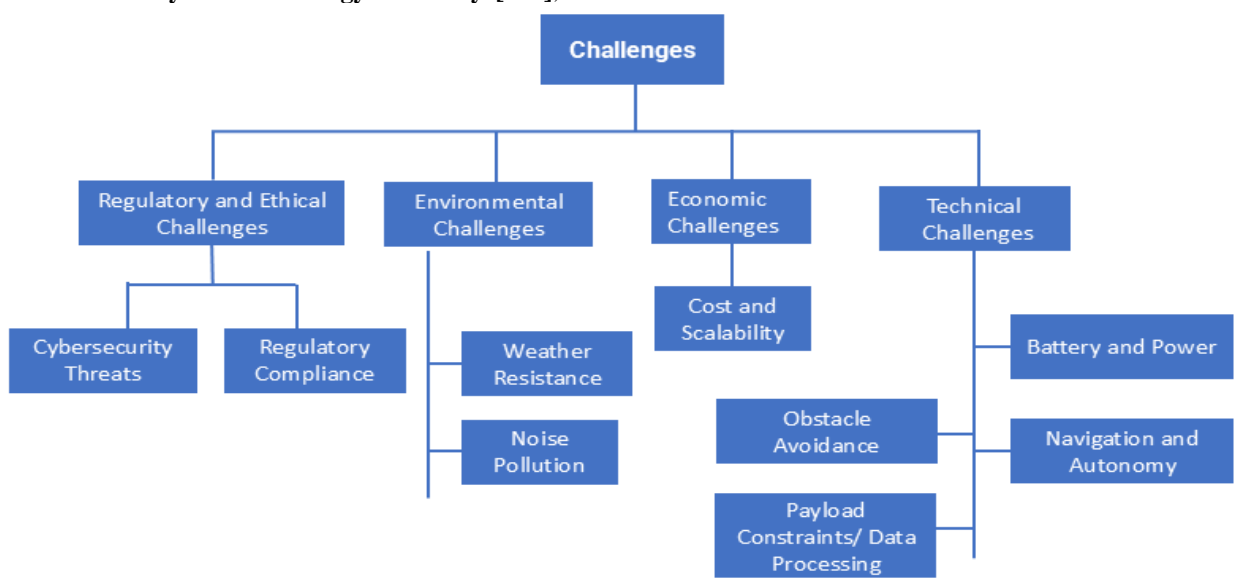

*Figure 21: Major Classification of Key Challenges in Drone Technology [204][205][206]*

Lastly, **sensor accuracy** poses an issue, as imperfections in sensor data can result in navigation errors and reduced reliability, particularly in complex or challenging environments. These technical hurdles collectively hinder the full potential of drones and require continued research and development efforts to overcome.

### 7.2 Regulatory and Ethical Challenges

Regulatory, ethical, and safety concerns present critical challenges to the widespread adoption of drones. Airspace integration is a significant issue, as harmonizing drone operations with existing manned aircraft systems requires careful planning and coordination. Additionally, compliance with diverse and evolving regulations across regions adds complexity for manufacturers and operators. From an ethical perspective, surveillance and privacy concerns arise due to unauthorized data collection by drones, while the potential for misuse, such as smuggling or espionage, amplifies security risks. Cybersecurity and safety concerns further complicate drone operations, with vulnerabilities to cyberattacks threatening control and data integrity. Additionally, collision risks, especially in complex environments with other drones or manned aircraft, underscore the need for advanced safety mechanisms. Addressing these challenges is essential to ensuring the responsible and secure deployment of drone technology. Regulatory and Ethical Challenges were discussed in [204][206][209]

### 7.3 Environmental Challenges

Environmental challenges pose significant barriers to the seamless integration of drones into various settings. **Noise pollution** from drone operations disrupts both urban environments and natural habitats, causing discomfort to humans and potential disturbances to the surrounding ecosystem. Additionally, drones can adversely affect **wildlife behavior**, leading to disruptions in natural ecosystems and posing risks to biodiversity.

These environmental impacts necessitate careful consideration and mitigation strategies to minimize harm and ensure sustainable drone usage. Environmental Challenges were discussed in [204][206]

### 7.4 Economic Challenges

Economic and cost factors present significant challenges to the widespread adoption of drones. The **high development costs** associated with researching and creating advanced drone technologies require substantial financial investment, making it difficult for smaller entities to compete in the market.

Additionally, the **affordability** of drones remains a concern, particularly in developing regions where limited resources and financial constraints hinder accessibility and adoption on a broader scale. These economic barriers emphasize the need for cost-effective solutions to promote the equitable integration of drones across various sectors. Economic Challenges were discussed in [204][206]

### 7.5 Summary Table

Table 21 illustrates the summary of Challenges in drone technology with their research highlights.

*Table 21: Summary of Challenges in Drone Technology [115][205][206][207][208]*

| Challenge | Description | Research Highlight |
|---|---|---|
| **Battery Life and Power** | Limited battery capacity restricts flight time and operational range. | Optimization of energy consumption and alternative energy sources are key focuses. |
| **Navigation Accuracy** | Achieving precise positioning in GPS-denied or dynamic environments. | Emphasis on SLAM, INS, and AI-based approaches for real-time localization. |
| **Obstacle Avoidance** | Ensuring real-time collision prevention in complex, unpredictable environments. | Development of LiDAR, ultrasonic sensors, and AI-based vision systems. |
| **Payload Capacity** | Balancing payload with operational efficiency and maneuverability. | Innovations in lightweight materials and modular designs. |
| **Regulatory Compliance** | Varied international and local regulations regarding drone operation, safety, and privacy. | Increasing focus on standardizing protocols and compliance frameworks. |
| **Weather Resistance** | Sensitivity to adverse weather conditions like wind, rain, and extreme temperatures. | Advances in aerodynamic designs and weather-resilient materials. |
| **Communication Reliability** | Maintaining robust and secure data links in high-interference environments. | Development of 5G-enabled communication systems and secure encryption protocols. |
| **AI and Data Processing** | Integrating real-time AI analytics for navigation, mapping, and decision-making. | Leveraging edge computing and AI for faster data processing and reduced latency. |
| **Cost of Deployment** | High cost of advanced UAV technologies limits accessibility for small businesses. | Focus on cost-efficient materials and scalable manufacturing techniques. |
| **Ethical Concerns** | Issues related to surveillance, privacy, and the potential misuse of drones for malicious activities. | Implementation of ethical frameworks and privacy protection mechanisms. |
| **Autonomy and Scalability** | Balancing full autonomy with scalable drone operations in fleet management. | Research on swarming algorithms and multi-agent coordination systems. |
| **Air Traffic Management** | Integrating drones into civilian airspace poses regulatory and technical challenges, including collision avoidance and traffic coordination. | Development of air traffic control systems and collision-prevention mechanisms. |
| **Environmental Factors** | Weather conditions such as strong winds, rain, and extreme temperatures significantly affect drone performance and reliability. | Implementation of weather-resilient materials and aerodynamic adaptations. |
| **Regulatory Constraints** | Global differences in drone regulations create barriers for international operations and scalability. | Efforts to harmonize regulations for global UAV deployments. |
| **Limited Autonomy** | Challenges in developing fully autonomous drones for complex, dynamic environments. | Research on swarming algorithms and multi-agent coordination systems. |
| **Noise Pollution** | Drones generate significant noise, which is a concern in urban and residential areas. | Innovations in noise-reduction technologies for quieter drone operations. |
| **Cybersecurity Threats** | Drones are vulnerable to hacking, data breaches, and GPS spoofing, compromising operations. | Implementation of encryption protocols and secure communication technologies. |

## 8. FUTURE TRENDS IN DRONE TECHNOLOGY

Future trends in drone technology encompass advancements in various domains, reflecting a blend of cutting-edge research and evolving real-world applications.

### 8.1 Swarm Intelligence and Multi-Agent Systems

Swarm intelligence and multi-agent systems represent a cutting-edge area of drone technology, where drones operate collectively by mimicking natural behaviors, such as flocks of birds. These systems have diverse applications, including **agriculture**, where they enable precision spraying of crops; **military operations**, such as coordinated reconnaissance; and **disaster response**, providing efficient collaboration in critical scenarios. Key innovations driving this field include the development of **swarming algorithms**, **decentralized decision-making** processes, and technologies that facilitate **real-time collaboration** among drones, enhancing their efficiency and effectiveness in complex tasks. Swarm Intelligence as a future trend were taken from [210].

### 8.2 Integration of Edge Computing and AI

The integration of edge computing and AI is transforming drone capabilities by enhancing onboard processing to minimize latency and optimize data handling. This advancement supports applications such as **real-time surveillance**, **autonomous navigation**, and **object recognition**, enabling drones to perform complex tasks with improved efficiency. Key technologies driving this integration include **neural networks**, which enable intelligent decision-making; **edge devices**, which provide localized processing power; and **federated learning**, which allows distributed AI training while maintaining data privacy and security. About Integration of Edge Computing and AI as a future trend were taken from [211][212][213].

### 8.3 Advanced Autonomy

Advanced autonomy is paving the way for drones capable of operating with minimal human intervention, revolutionizing various industries. These fully autonomous drones find applications in **logistics** for package delivery, **urban management**, and **industrial monitoring**, offering efficiency and convenience. However, achieving this level of autonomy presents challenges, including addressing **ethical concerns**, ensuring robust **collision avoidance** mechanisms, and managing the **scalability** of autonomous systems for widespread deployment [131].

### 8.4 Regulatory Innovations

Regulatory innovations are essential for addressing global disparities in drone regulations, enabling smoother integration of drones into commercial and public domains. Key focus areas include the development of **standardized air traffic management systems** to ensure safe and efficient operations, the implementation of **dynamic licensing frameworks** to accommodate diverse use cases, and the establishment of robust **privacy regulations** to protect individuals and communities from potential misuse. About Regulatory Innovations as a future trend were taken from [204][206].

### 8.5 Green Energy Solutions

The transition to sustainable energy sources is a pivotal step in reducing the environmental impact of drones.

Key technologies driving this shift include **solar-powered UAVs**, which harness renewable energy for extended operations, **hydrogen fuel cells**, offering clean and efficient power solutions, and **energy-efficient designs**, aimed at optimizing performance while minimizing resource consumption [214][215][216].

### 8.6 Enhanced Applications in Urban Infrastructure

Enhanced applications in urban infrastructure are expanding the role of UAVs in smart cities, enabling tasks such as real-time traffic monitoring and emergency services. These advancements are driven by key features including **IoT integration** for seamless connectivity, **predictive maintenance** to ensure operational efficiency, and **AI-based urban planning** to support intelligent decision-making and resource management [217].

### 8.7 Augmented Reality (AR) and Immersive Experiences

Augmented Reality (AR) is being integrated into drone technology to revolutionize media, tourism, and educational applications. Examples include drone-assisted AR games, which provide immersive experiences for users, virtual heritage tours that allow remote exploration of historical sites, and live-event broadcasting, where drones capture dynamic footage for enhanced viewer engagement. These innovations are transforming how people interact with digital content in real-time.[218][219][220][221]

### 8.8 Emergence of Drone-as-a-Service (DaaS)

The emergence of Drone-as-a-Service (DaaS) is commercializing drone operations, offering on-demand services for businesses across various industries. This model is particularly impactful in sectors like e-commerce, where drones facilitate rapid delivery services, healthcare, where they are used for transporting medical supplies, and agriculture, where drones assist in precision farming and monitoring crop health. By offering flexible, scalable solutions, DaaS is reshaping how industries approach logistics and operations [222].

### 8.9 Nano and Micro-Drones

The development of smaller drones is pushing the boundaries of drone technology, enabling their use in niche applications. These compact drones are particularly valuable in fields such as medical diagnostics, where they can assist in remote health monitoring, surveillance in tight spaces for security and search operations, and precision farming, where they enable detailed monitoring of crops and soil conditions. Their small size allows for greater maneuverability and access to areas that larger drones cannot reach, expanding their potential across various sectors.[223][224][225]

### 8.10. Integration with 5G Networks

The integration of 5G technology into drone operations leverages its low latency and high bandwidth to significantly enhance connectivity. This advancement enables real-time analytics, allowing drones to process and transmit data more efficiently. It also facilitates remote control, enhancing the operation of unmanned aerial vehicles (UAVs) over long distances. Additionally, 5G supports the seamless coordination of UAV fleets, ensuring smooth and reliable operations, particularly in complex, data-intensive environments such as industrial inspections and autonomous delivery systems [222].

## 9. DRONE APPLICATION CASE STUDY

### 9.1 Disaster Management in Turkey (2023 Earthquake)

In February 2023, Turkey experienced a devastating earthquake that caused widespread destruction, particularly in urban areas, severely hampering traditional search and rescue operations. Collapsed infrastructure delayed access to survivors, highlighting the critical need for innovative solutions to address these challenges. Fig. 22 shows the images captured by drone.UAVs (unmanned aerial vehicles) played a transformative role in disaster management during this crisis. Equipped with advanced thermal cameras, these drones effectively detected heat signatures of survivors trapped under rubble, even in low visibility or nighttime conditions. AI-powered object recognition systems enhanced efficiency by distinguishing human survivors from other objects, while SLAM (Simultaneous Localization and Mapping) technology provided real-time mapping of the affected areas, ensuring precise navigation and situational awareness for rescue teams. Table 22. gives summary of Disaster Management in Turkey [226].

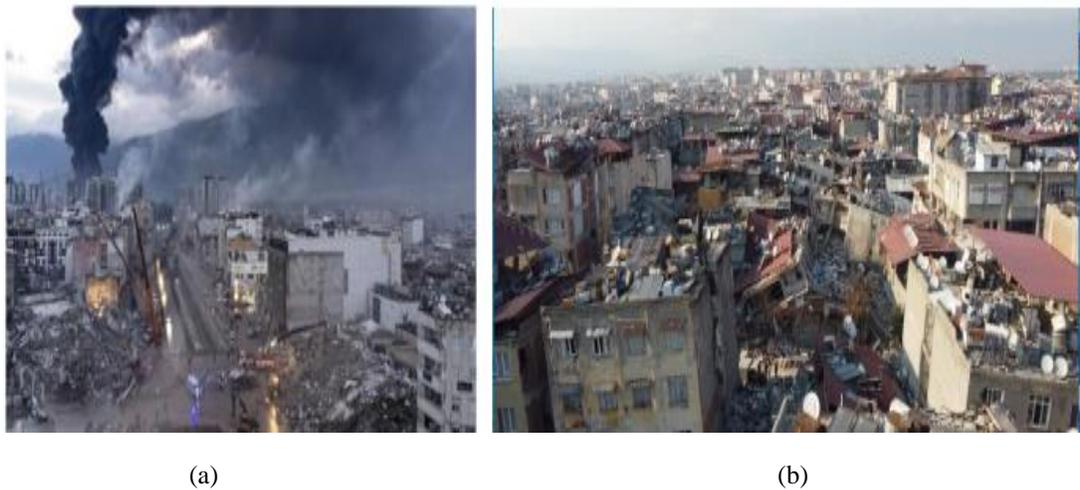

(a)                                                    (b)

*Figure 22. Drone Captured Images (a) Collapses in Iskenderun on both sides of the road; the amount of obstruction is much larger than a 45° angle may suggest. (b) Partial building dismantled in Hatay, requiring total demolition (with copyright permission from PEMA [226][227]*

### 9.2 Amazon Prime Air (San Francisco) [228]

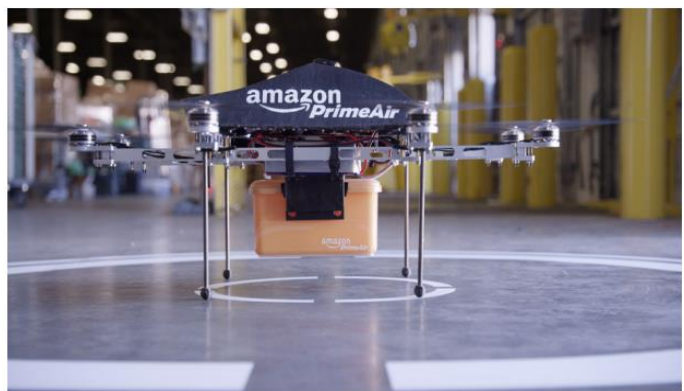

*Figure 23: Drone-based Delivery of Package [230]*

This case study examines the Amazon Prime Air initiative, a drone-based delivery system that has been operational since 2017 in regions such as San Francisco, USA. By leveraging advanced technologies like GPS navigation, AI optimization, and obstacle avoidance, this initiative aims to address inefficiencies in last-mile logistics and promote sustainable delivery solutions. Fig 23 shows the image of drone used for package delivery. Information about this case study was referenced from [228][229].

### 9.3 Wildlife Monitoring in Rainforest

The study focuses on improving data precision and wildlife monitoring efficiency through high-resolution imaging and thermal mapping. By leveraging UAVs, the researchers addressed challenges such as detecting wildlife in inaccessible areas, reducing human disturbance, and collecting real-time, actionable data. This approach also facilitates the creation of Ortho maps for habitat analysis and aids in conservation efforts. Information about this case study was referenced from [231]

### 9.4 Zipline Drones for Healthcare (Rwanda)

"The Blood is Here: Zipline's Medical Delivery Drones Are Changing the Game in Rwanda" explores Zipline's innovative use of fixed-wing drones to deliver critical medical supplies, such as blood and medications, to remote areas in Rwanda. Fig. 24 shows how zipline drones delivers blood to the hospital in Rwanda.

Operating since 2016, these drones bypass challenging terrains and slow transportation infrastructure, ensuring faster and more reliable deliveries. The service, which significantly reduces delivery times, has saved countless lives, particularly in emergencies. The system also demonstrates scalability and reliability for wider applications, including disaster responses and military scenarios [232][233][234].

### 9.5 Traffic Monitoring in Dubai

Dubai's smart city vision incorporates UAVs to revolutionize urban management, addressing challenges like traffic congestion, emergency response, and public safety. This case study explores the implementation of drone technology and its implications for future urban planning [235][236][237].

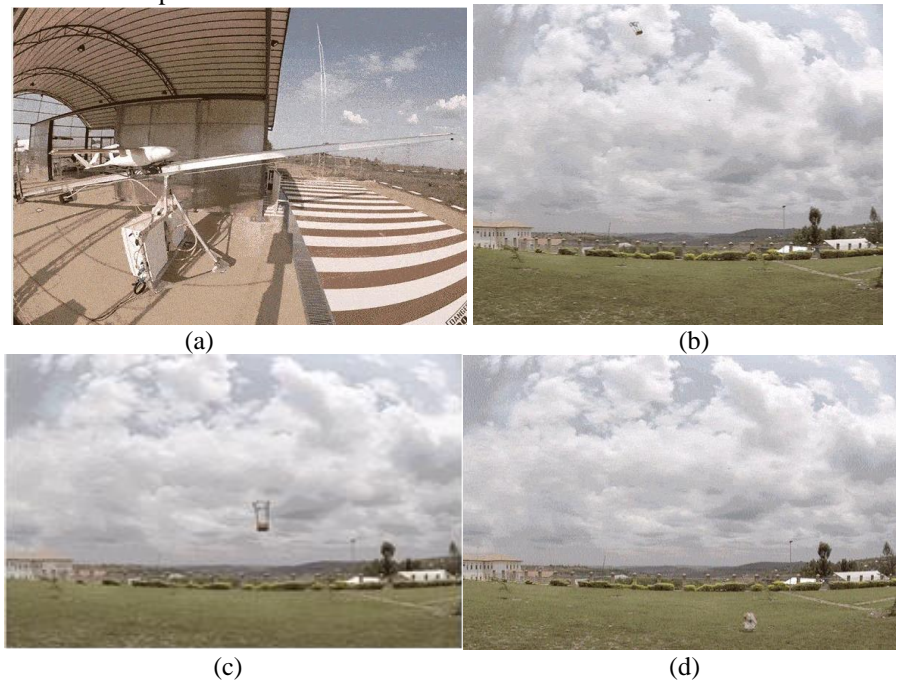

(a)          (b)

(c)          (d)

Figure 24: Images of Ziplines Drone Delivering Blood. (a) Drone Take-off to delivery Blood to hospital. (b) A view of drone with parachute to deliver package. (c) A view of drone path in between the destined location. (d) Image of package delivered to location [232]

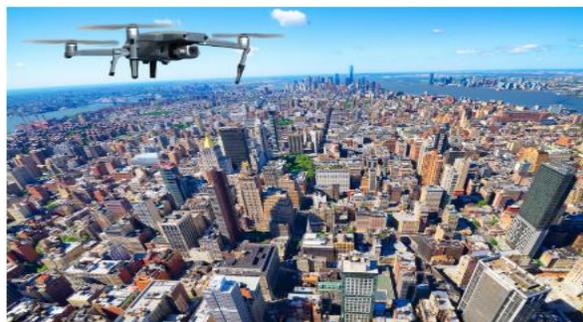

Figure 25: NYC Department of Buildings Examines Drone use for Inspections [238]

*Table 22: Summary table of case study [226] ... [ 244]*

| S.No | Case Study | Context | Technologies | Drone Type/Model Used | Year | Issues Resolved | Impact |
|---|---|---|---|---|---|---|---|
| 1 | Disaster Management in Turkey. (2023 Earthquake) | Drones aided search and rescue operations post-earthquake. | Thermal cameras, AI object recognition, SLAM | DJI Matrice 300, Autel EVO II Dual | February 2023 | Delayed access to survivors due to collapsed infrastructure. | Reduced response times, improved survivor detection. |
| 2 | Amazon Prime Air. (San Francisco) | Drone-based delivery of packages. | GPS navigation, AI optimization, obstacle avoidance | Amazon Prime Air Drones | Ongoing since 2017 | Inefficient last-mile logistics and environmental concerns. | Reduced delivery times, logistics cost, eco-friendly operations |
| 3 | Wildlife Monitoring in Rainforest | Conservation drones tracked biodiversity and detected illegal logging. | Multispectral sensors, AI pattern analysis, real-time streaming | Parrot Anafi, eBee X | Ongoing since 2021 | Biodiversity loss and challenges in monitoring deforestation. | Enhanced wildlife protection and deforestation control. |
| 4 | Zipline Drones for Healthcare (Rwanda). | Delivered blood and vaccines to remote clinics. | IoT-enabled tracking, GPS routing, real-time analytics | Zipline Drone | Ongoing since 2016 | Inaccessible rural areas delaying critical medical supplies. | Addressed medical supply delays, saving thousands of lives. |
| 5 | High-Rise Building Inspections (New York) | Inspected skyscrapers for structural integrity. | LiDAR, photogrammetry, AI for crack detection | DJI Matrice 210 RTK | March 2024 | Dangerous and costly manual inspections of high-rise structures | Reduced inspection costs, increased worker safety. |
| 6 | Traffic Monitoring in Dubai. | Analyzed urban congestion for smart city planning. | AI-based traffic analysis, IoT sensors, real-time data sharing | DJI Phantom 4 Pro | January 2023 | Urban traffic congestion and lack of real-time traffic data. | Improved urban planning, reduced congestion. |
| 7 | Inspection of Energy Infrastructure. (Texas, Abu Dhabi, Guangzhou China,) | Monitored wind turbine, hydrocarbon gas pipeline and solar farms. | Infrared imaging, edge computing, real-time reporting | DJI Matrice 300 RTK | December 2022 | Difficulties in detecting energy system faults efficiently. | Improved efficiency, early fault detection. |
| 8 | Precision Agriculture. (Japan) | Sprayed pesticides and monitored rice fields. | Multispectral imaging, automated spraying systems | Terra Drone | Ongoing since 2020 | Overuse of pesticides and lack of precision farming data. | Reduced chemical use, increased yield. |
| 9 | Kerala Floods. (India, 2018) | Surveyed flood-affected areas for rescue efforts. | High-resolution cameras, real-time mapping, GPS | DJI Inspire 2, DJI Mavic 2 Pro | August 2018 | Lack of real-time flood mapping for rescue and relief planning. | Enhanced relief planning and resource allocation. |
| 10 | Port Surveillance. (US) | Monitored port activities for logistics and security. | Thermal cameras, AI analysis, night vision | DJI Matrice 210 | Ongoing since 2019 | Security breaches and inefficiencies in cargo management. | Improved cargo management, theft prevention. |

By including these recent and diverse studies, the Table 22. not only highlights the latest advancements in drone technology but also connects them to real-world implementations, making the references both current and highly relevant for ongoing research

### 9.6 Inspection of Energy Infrastructure [239]

#### 9.6.1 Internal inspection of wind turbine blade
Wind farms, often located in remote and difficult-to-access areas, require regular inspections to ensure optimal operation. Traditional inspections are slow and dangerous, as technicians must climb the turbine. The Danish power company Orsted tested the Elios 2 for an inspection of an offshore wind turbine blade and found that it helped them inspect 40% of the blade length, improve safety, and halve the time required for the inspection[1]. Fig. 26 (a) shows the inside image of the blade taken from the Elios 2 drone. Today the new Elios 3 model incorporates many improvements, enhancing the use of the inspection drone in many confined space applications with high image resolution and scanning power.

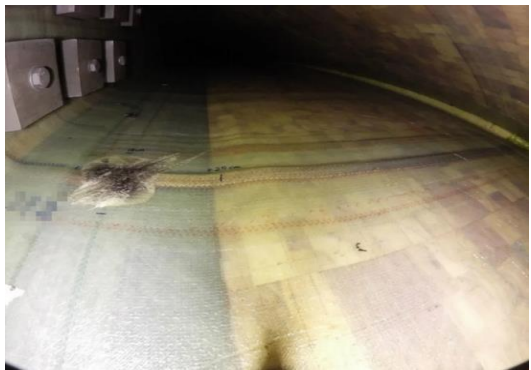
(a)

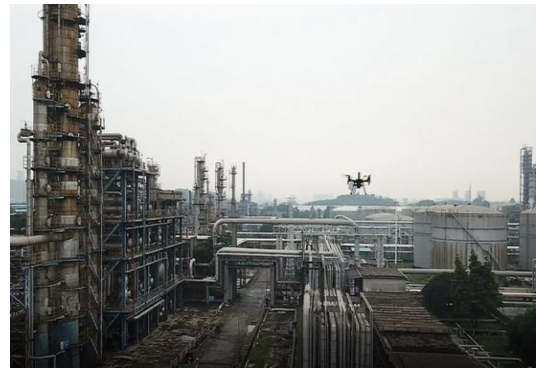
(b)

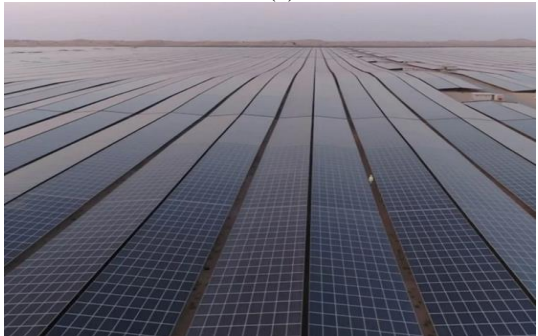
(c)

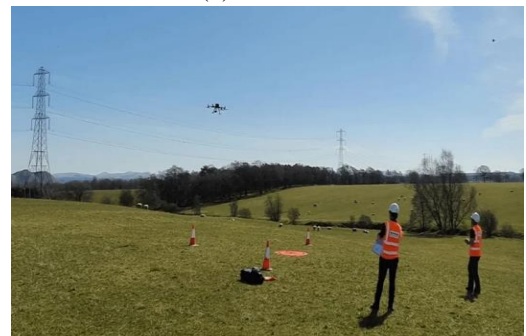
(d)

*Figure 26: Inspection by drone (a) Inspection of wind turbine blades [239]. (b) Drone inspection of hydrocarbon gas piping systems for leak detection [239]. (c) Inspection of solar power plant.[239] (d) Inspection of transmission towers [239].*

#### 9.6.2. Drone inspection of hydrocarbon gas piping systems for leak detection
For companies in the oil and gas sector, an important part of their safety work consists of monitoring hydrocarbon gas service pipeline systems for leak detection and correction. However, due to the wide and complex distribution of pipelines, traditional manual inspection methods present problems of performance, personnel risk, and accessibility, among others.

At a petrochemical complex in Guangzhou China, Soarability provided a suite of Sniffer4D gas detection systems designed specifically for pipeline inspection and mounted on Matrice 350 RTK drones. At the site, the survey team operated the drone in such a way that it flew and scanned the area within a 3-meter radius above the pipelines. Meanwhile, they paid close attention to the concentrations of hydrocarbon gases, mainly methane, and the gradient changes shown by using Sniffer4D Mapper[2] software. Figure 26 (b) shows a photograph of the Matrice 350 RTK drone equipped with the Sniffer4D thermal camera, flying over oil and gas pipelines.

#### 9.6.3 Drone inspection of solar power plant
Solar plants require regular inspections to ensure that the solar panels are in good condition and operating at maximum capacity. Manual inspections are tedious and time-consuming. Inconsistent or inaccurate inspections can cause damage, which in turn leads to lost efficiency and profits. In the case of solar, this is even more critical, as there are large communities that rely on them to operate smoothly.

One location for solar power plants is the Middle East, where large desert areas with no residents are ideal for solar panel farms. In addition, the almost perpetually clear skies allow for constant power generation.

Noor Abu Dhabi, located near the city of Sweihan, is 8 square kilometers in size and has more than 3.2 million solar panels. It is currently the largest solar power plant in the world located at a single site.

Pix4D's distributor and partner, Falcon Eye Drones (FEDS), was contracted to track and monitor monthly progress during the plant's construction with accurate, high-resolution drone mapping. The desired results included maps, videos, and individual images[3] (figure 26 (c)).

### 9.6.4 Drone inspection of transmission towers

Transmission towers, which transport electricity over long distances, require regular inspections to prevent failures. Traditional inspections are dangerous and time-consuming, and in other cases, to overcome this low efficiency, the use of helicopters is required, which substantially increases the cost of the inspection. Fig 26 (d) shows drone inspection of transmission power.

An electric utility company began using drones equipped with high-resolution cameras and infrared sensors to inspect its transmission towers. The drones can fly close to the towers and capture detailed images of critical components.

Leading engineering firm Keltbray, a specialist provider of overhead power line and substation services, is using drones to double inspection efficiency and reduce carbon emissions by nearly 50%.

Keltbray has invested in the DJI M300 RTK platform and its ecosystem of powerful cameras to collect higher-quality data, faster and more securely than ever before.

The P1 photogrammetry camera and L1 LiDAR sensor are integrated into the drone to build highly detailed maps, models and point clouds to improve mission planning and address the challenges associated with traditional power line construction methods[4].

The H20T high-resolution thermal imaging camera is also integrated into the vehicle to perform thermal evaluations of power lines for early detection of defects or thermal anomalies that could compromise the continuity of power supply.

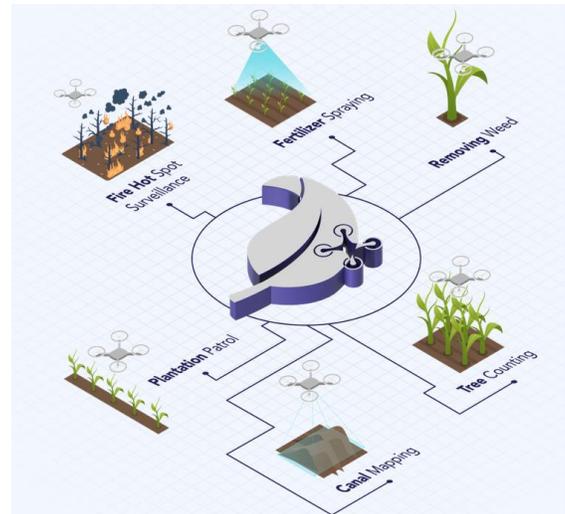

*Figure 27: Precision Agriculture in Japan by Drone [244].*

### 9.7 Kerala Floods (India, 2018)

The 2018 Kerala floods, the most severe in a century, displaced over 314,000 individuals and severely disrupted transportation and communication networks. Traditional rescue methods faced challenges, with helicopters limited by deployment difficulties and high altitudes. Drones emerged as a critical tool, capable of low-altitude flights for real-time monitoring, damage assessment, and locating stranded individuals on rooftops. Fig. 28 gives images of flood taken by drone Easily deployable in large numbers, UAVs offered flexibility and efficiency in inaccessible areas. This response underscored the potential of drones in disaster management and emergency planning worldwide [240][241].

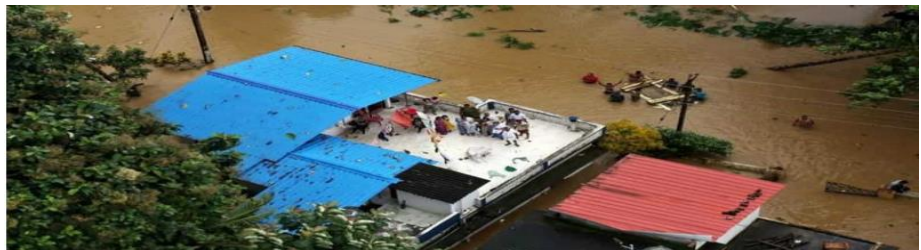

*Figure 28: Kerala flood Images Captured by Drone [241].*

## Novelty of Survey Paper

This paper distinguishes itself by offering an integrative and comprehensive review of the drone landscape, addressing gaps in existing literature. Unlike previous studies that focuses on aspects such as applications or navigation techniques. The current survey unifies historical perspectives, classification, architectural insights, navigation advancements, and future trends along with key challenges of the technology. By systematically analyzing over 240 papers and adopting the PRISMA methodology, this work establishes a consolidated resource for researchers. The inclusion of case studies and practical insights further enhances its utility, providing real-world contexts that bridge theoretical frameworks with actionable knowledge. More-over the drone classification framework presented in this paper represents a **novel contribution to the field of unmanned aerial systems (UAS)** by integrating a wide array of classification criteria into a unified, multi-dimensional framework. The novelty of this framework lies in its ability to comprehensively categorize drones based on diverse attributes, bridging the gap between design, functionality, and application-specific requirements. The layered architectural design of drone discussed in this paper not only enhances scalability and interoperability but also supports innovation by allowing upgrades or integration of new technologies in specific layers without disrupting the entire system. Such a comprehensive architecture is crucial for meeting the diverse requirements of modern drone applications across industries.

## Conclusion

Drones development have become more advanced with the advent of new technology such as AI, navigation system, lightweight materials from simple military tool to be used in every field of life. This survey discusses a few of the transformative opportunities offered with drone technology especially in agriculture, logistics, health-care, and disaster management along with notable challenges such as regulatory constraints, cybersecurity threats and technical limitations. The paper highlights the need for continued collaboration and research in order to overcome these barriers and unlock the potential of drones. The Future of Drones The drone of 2024 will have increased capabilities, autonomy and long-term, with the imminent supply of drone ecology, drone transport and autonomous delivery The research has been foundational and provides the basis for subsequent work, while also enabling innovation in drone technologies. This work serves as a foundational reference, inspiring future research and fostering innovation in drone technology.